\documentclass[preprint,12pt]{elsarticle}

\usepackage{amsmath,amssymb}
\usepackage{graphicx}
\usepackage{subfigure}
\usepackage{setspace}
\usepackage{float}
\usepackage{graphicx}
\usepackage{epsfig,epstopdf}
\usepackage{verbatim}
\usepackage[section]{algorithm}
\usepackage{algorithmic}
\usepackage{amsmath,amssymb}
\usepackage{graphicx}
\usepackage{subfigure}
\usepackage{setspace}
\usepackage{float}
\usepackage{graphicx}
\usepackage{epsfig,epstopdf}
\usepackage{verbatim}
\usepackage{color}
\usepackage[section]{algorithm}
\usepackage{algorithmic}
\usepackage{array}
\usepackage{amsmath}
\usepackage{amssymb}
\usepackage{amsfonts}
\usepackage{amsthm}
\usepackage{booktabs}
\usepackage{color}
\usepackage{graphicx}
\usepackage{subfigure}
\usepackage{multirow}
\usepackage{threeparttable}
\usepackage{hyperref}
\usepackage{makecell}
\usepackage{fmtcount}



\newcolumntype{L}[1]{>{\raggedright\let\newline\\\arraybackslash\hspace{0pt}}m{#1}}
\newcolumntype{C}[1]{>{\centering\let\newline\\\arraybackslash\hspace{0pt}}m{#1}}
\newcolumntype{R}[1]{>{\raggedleft\let\newline\\\arraybackslash\hspace{0pt}}m{#1}}

\numberwithin{equation}{section}

\theoremstyle{remark}

\newcommand{\mat}[1]{\ensuremath{\mathbf{#1}}}

\newcommand{\cG}{\mathcal{G}}

\newcommand{\eg}{e.g.}
\newcommand{\ie}{i.e.}

\newcommand{\cL}{\mathcal{L}}
\newcommand{\cD}{\mathcal{D}}

\newcommand{\R}{\mathbb{R}}

\newcommand{\scal}[2]{\left\langle #1,#2 \right\rangle}
\newcommand{\la}{\langle}
\newcommand{\ra}{\rangle}

\newcommand{\suml}[2]{\sum\limits_{#1}^{#2}}

\begin{document}

\begin{frontmatter}
\journal{Journal of Mathematical Imaging and Vision}

\title{Speckle Reduction with Trained Nonlinear Diffusion Filtering}
%
%
%

\author{Wensen Feng\fnref{label1}}
\author[label3]{Yunjin Chen \corref{cor1}}
\ead{chenyunjin\_nudt@hotmail.com}
\cortext[cor1]{Corresponding author: Yunjin Chen}
\address[label1]{School of Automation and Electrical Engineering, University of Science and Technology Beijing, Beijing,
100083, China.}
\address[label3]{Institute for Computer Graphics and Vision, Graz University of Technology, Inffeldgasse 16, A-8010 Graz, Austria.}

\address{}

\begin{abstract}
Speckle reduction is a prerequisite for many image processing tasks in synthetic
aperture radar (SAR) images, as well as all coherent images.
In recent years, predominant state-of-the-art approaches for despeckling are
usually based on nonlocal methods which mainly concentrate on achieving utmost
image restoration quality, with relatively low computational efficiency.
Therefore, in this study we aim to propose an efficient despeckling model with
both high computational efficiency and high recovery quality.
To this end, we exploit a newly-developed trainable nonlinear reaction diffusion(TNRD)
framework which has proven a simple and effective model for various image
restoration problems. {In the original TNRD applications,
the diffusion network is usually derived based on
the direct gradient descent scheme. However, this approach will encounter some
problem for the task of multiplicative noise reduction exploited in this study. To solve this problem, we employed a new architecture derived from the proximal gradient descent method.} {Taking into account the speckle noise statistics,
the diffusion process for the despeckling task is derived. We then retrain all
the model parameters in the presence of speckle noise. Finally,
optimized nonlinear diffusion filtering models are obtained, which are specialized
for despeckling with various noise levels.
Experimental results substantiate that the trained filtering models provide comparable
or even better results than state-of-the-art nonlocal approaches.
Meanwhile, our proposed model merely contains convolution of linear filters with
an image, which offers high level parallelism on GPUs. As a consequence,
for images of size $512 \times 512$, our GPU implementation takes less than 0.1
seconds to produce state-of-the-art despeckling performance.}
\end{abstract}


\end{frontmatter}


\section{Introduction}

Synthetic aperture radar (SAR) images are inevitably corrupted by
speckle noise, due to constructive and destructive electromagnetic wave
interference during image acquisition. With fleets of satellites delivering a huge number
of images, automatic analysis tools are essential for remote sensing major applications.
Therefore, the quality of source images should be sufficient such that it is
easy to extract information. However, the speckle noise visually degrades the
appearance of images and therefore hinders automatic
scene analysis and information extraction \cite{argenti2013tutorial, di2014benchmarking}.
For instance, speckle is the main
obstacle towards the development of an effective optical-SAR fusion
\cite{alparone2004landsat}. {
Hence, speckle reduction is a necessary preprocessing step in SAR image processing.
Despeckling techniques have been extensively studied for almost 30
years \cite{lee1980digital, kuan1987adaptive}, and new algorithms are continuously
proposed to provide better and better performance.}
Up to now, the despeckling techniques fall broadly into four categories:
filtering based methods in (1) spatial domain or (2) a transform domain, \eg,
wavelet domain; (3) nonlocal filtering; and
(4) variational methods.
{As a comprehensive review of the despeckling algorithms is
beyond the scope of this paper, we only provide a brief introduction for these methods.
For more details, we refer the reader to \cite{argenti2013tutorial}.}

The multi-look technique is a traditional spatial approach. It amounts to incoherently averaging independent observations of the same resolution cell, thus reducing the noise intensity. However, this simple averaging approach results in a clear loss in image resolution. To overcome this deficiency, a great deal of research has been conducted to develop suitable spatial filters which can reduce the noise, yet preserve details and
edges \cite{Frostpami1982} \cite{lopes1993structure}. Filters of this kind include
Lee filter proposed in \cite{lee1980digital} \cite{lee1981refined} which was developed under
the minimum-mean-square-error (MMSE) criterion, and Kuan filter \cite{kuan1987adaptive}
as well as the $\Gamma$-Map filter \cite{lopes1990maximum} which are based on the
more sophisticated maximum a posteriori (MAP) approach.

{
Anisotropic diffusion \cite{perona1990scale} based method is also a type of widely
exploited spatial filtering technology for despeckling.
Anisotropic diffusion is a popular technique
in the image processing community, that aims at reducing image noise without
removing significant parts of the image content. A few related works that
apply anisotropic diffusion filtering for the despeckling task include speckle
reducing anisotropic diffusion (SRAD) \cite{yu2002speckle} and
detail preserving anisotropic diffusion (DPAD) \cite{aja2006estimation}.
SRAD exploits the instantaneous coefficient of variation and it
leads to better performance than the conventional anisotropic diffusion
method in terms of mean preservation, variance reduction and edge
localization. DPAD modifies the SRAD filter to rely on the Kuan filter
\cite{kuan1987adaptive} rather than the Lee filter. DPAD estimates the local
statistics using a larger neighborhood, instead of the four direct neighbors used by
SRAD. However, the despeckling methods based on anisotropic diffusion fell out of
favor in recent years mainly because of limited performance. It is clear that
there is a despeckling quality gap between the diffusion based approaches
and state-of-the-art nonlocal algorithms.}

Image filtering in the domain of wavelet has also been widely exploited for
despeckling. Most of the wavelet-based despeckling techniques employ the statistical
wavelet shrinkage technique with MAP Bayesian approach, \eg
\cite{RanjaniTGRS, achim2006sar, bianchi2008segmentation,
ranjani2011generalized, argenti2012fast}.
In general, the wavelet-based methods guarantee a superior ability to preserve signal
resolution in comparison with conventional spatial filters. However, they often suffer from
isolated patterns in flat areas, or ringing effects near the edges of the images, leading to
visually unappealing results.

Recently, incorporation with the modern denoising methods, \eg, nonlocal mean (NLM)
\cite{buades2005review}, block-matching 3-D (BM3D) \cite{dabov2007image} and K-SVD
\cite{elad2006image}, several nonlocal despeckling approaches have been proposed
\cite{deledalle2009iterative}, \cite{teuber2012nonlocal}, \cite{ParrilliPAV12},
\cite{huang2012multiplicative}.
Originated from the NLM algorithm, the probabilistic patch-based (PPB) filter
\cite{deledalle2009iterative} provides promising results by developing an
effective similarity measure well suited to SAR images.
A drawback of the PPB filter is the suppression of thin and dark details in the
regularized images. As an extension of BM3D, Parrilli $et$ $al.$ \cite{ParrilliPAV12}
derived a SAR-oriented version of BM3D by taking into account the peculiar features of
SAR images. It exhibits an objective performance comparable or superior to other
techniques on simulated speckled images, and guarantees a very good subjective quality
on real SAR images. Typical artifacts of the nonlocal methods are in the form of
structured signal-like patches in flat areas, originated from the randomness
of speckle and reinforced through the patch selection process.
Generally speaking, most of these techniques mainly concentrate on achieving
utmost despeckling quality, with relatively low computational efficiency.
An notable exception is BM3D with its improved version \cite{CozzolinoPSPV14}.
However, the BM3D-based method involves a block matching process, which is
challenging for parallel computation on GPUs, alluding to the
fact that it is not straightforward to accelerate BM3D algorithm
on parallel architectures.

The fourth category, i.e., variational methods \cite{SteidlT10,
AubertA08, wensen14, chen2014higher, durand2010multiplicative, jin2011variational, chen2016multiplicative, escandevariational,zengtie1,zengtie2}, minimizes some appropriate energy
functionals consisted of
an image prior regularizer and a data fitting term. As a well-known regularizer,
total variation (TV) has been widely used for the despeckling task \cite{SteidlT10,
AubertA08,zengtie1,zengtie2}. {For instance, in \cite{zengtie1}
a new variational model based on a hybrid data term and the widely
used TV regularizer is proposed for restoring blurred images with multiplicative noise. Moreover, \cite{zengtie2} proposes a two-step approach to solve the problem of restoring images degraded by multiplicative noise and blurring, where the multiplicative noise
is first reduced by nonlocal filters and then a convex variational model is
adopted to obtain the final restored images.} Solutions of variational problems with TV regularization admit
many desirable properties, most notably the appearance of sharp edges.

{
However, TV-based methods generate the so-called staircasing artifact.
To remedy the staircasing artifact, \cite{wensen14} incorporates the total
generalized variation (TGV) \cite{BrediesKP10} penalty into the existing data
fidelity term for speckle removal, and develops two novel variation despeckling models.
By involving and balancing higher-order derivatives of the image, the TGV-based
despeckling method outperforms the traditional TV methods by reducing the
staircasing artifact. Recently, different from hand-crafted regularizers,
such as TV and TGV models, \cite{chen2014higher} proposes a novel variational approach
for speckle removal, which combines an image prior model named
Fields of Experts (FoE) \cite{RothFOE2009}} and a recently proposed efficient
non-convex optimization algorithm - iPiano \cite{iPiano}.
The proposed method in \cite{chen2014higher} can obtain strongly competitive despeckling
performance w.r.t. the state-of-the-art method - SAR-BM3D,
meanwhile, preserve the property of computational efficiency.

\subsection{Our Contribution}
Traditional despeckling approaches based on anisotropic diffusion are handcrafted
models which include elaborate selections of diffusivity coefficient,
or optimal stopping time or proper reaction force term.
In order to improve the capacity of the traditional diffusion-based despeckling models,
we employ the newly-developed anisotropic diffusion based model \cite{TNRD} with
trainable filters and influence functions.
Instead of the first order gradient operator in previous diffusion-based despeckling
models, we explore more filters of larger kernel size targeted for despeckling.
On the other hand, different influence functions are considered and trained for
different filters, rather than an unique function in the traditional diffusion model.
Moreover, the parameters of each iteration can vary across diffusion steps.

{
As shown in \cite{TNRD}, the optimized nonlinear diffusion model has broad applicability
to a variety of image restoration problems, and achieves recovery results of high
quality surpassing recent state-of-the-arts.
Furthermore, it only involves a small number of explicit filtering steps,
and hence is highly computationally efficient, especially with parallel
computation on GPUs.}

{
In this paper, we intend
to apply the TNRD framework \cite{TNRD} to the task of multiplicative noise reduction.
However, a direct use of the original TNRD model is not feasible, as we have to
make a few modifications oriented to this specific task:
\begin{itemize}
\item We need to redesign the diffusion process by taking into consideration
the peculiarity of multiplicative noise statistics.
\item Based on the new diffusion process specialized for multiplicative noise
reduction, we need to recalculate the gradients required for the training phase.
\item In the original TNRD applications,
the diffusion network is usually derived based on
the direct gradient descent scheme. However, this approach will encounter some
problem for the task of multiplicative noise reduction exploited in this work. The reason is explained in detail in Section \ref{despeckling_model} and the experimental part \ref{compareexperi}.
To solve this problem, we employed a new architecture as shown in Fig.~\ref{fig:feedforwardCNNspeckle},
which is derived from the proximal gradient descent
method. Comparing Fig.~\ref{fig:feedforwardCNNspeckle} and Fig.~\ref{fig:feedforwardCNN}, we can see that the structure of the proposed
diffusion process Fig.~\ref{fig:feedforwardCNNspeckle} in our study is quite different from the original TNRD model Fig.~\ref{fig:feedforwardCNN}.
\end{itemize}
%
Then, the model parameters in the diffusion process need to be trained by taking into account the Speckle noise statistics, including the linear filters and influence functions. Eventually, we reach a
nonlinear reaction diffusion based approach for despeckling, which leads to state-of-the-art performance, meanwhile gains high computationally efficiency.}
Experimental results show that the proposed despeckling approach with optimized
nonlinear diffusion filtering leads to state-of-the-art performance,
meanwhile gains high computationally efficiency.

\subsection{Organization}
The remainder of the paper is organized as follows. Section II presents
a general review of the speckle noise and the trainable nonlinear reaction diffusion
process, which is required to derive the optimized diffusion process for despeckling.
In the subsequent section III, we propose the optimized nonlinear diffusion
process for despeckling.
Subsequently, Section IV describes comprehensive experiment results for the proposed model.
The concluding remarks are drawn in the final Section V.
\section{Preliminaries}
\label{Preliminaries}
To make the paper self-contained, in this section we provide a brief review of
the statistics property of speckle noise and the trainable nonlinear diffusion
process proposed in \cite{TNRD}.

\subsection{Speckle Noise}
Assume that $f \in \mathbb{R}^N$ (represented as a column-stacked vector) denotes the observed SAR image amplitude with number of looks $L$, and $u \in \mathbb{R}^N$ denotes the underlying true image amplitude i.e., the square root of the reflectivity. According to \cite{goodman1975statistical} \cite{argenti2013tutorial}, the fully-developed speckle can be modeled as the multiplicative Goodman's noise, i.e.,
\begin{equation}
\label{mul}
f = un,
\end{equation}
where $n$ is the so-called speckle noise.
In the multiplicative, or fully-developed,
speckle model, the conditional pdf of $f$ given $u$ follows a Nakagami distribution:
\[
p(f \vert u) = \frac {2L^L}{\Gamma(L)u^{2L}}f^{2L-1}\text{exp}\left(-\frac {Lf^2}{u^2}\right) \,,
\]
where $\Gamma$ is the classical Gamma function. Note that the distribution
According to the Gibbs function, this likelihood leads to the
following energy term via $E = -\text{log}p(f \vert u)$
\begin{equation}\label{dataterm1}
D_1(u, f) = \langle L\cdot (2\text{log} u + \frac {f^2}{u^2} ),1 \rangle \,,
\end{equation}
where $\langle , \rangle$ denotes inner product.
Unfortunately, this term is nonconvex w.r.t. $u$, that is cumbersome to derive
the diffusion process as described in Sec. \ref{despeckling_model}.
There are two ways to resolve the nonconvexity.
First, logarithmic transformation (i.e., $u_l = \text{log} \left(u \right)$) can be used. The resulting energy term is expressed as follows:
\begin{equation}\label{dataterm2}
D_2(u_l, f) = \langle L\cdot (2u_l + {f^2}{e^{-2u_l}} ),1 \rangle \,.
\end{equation}
Second, an alternative method to define a convex date term is using
the classical Csisz{\'a}r I-divergence model \cite{csiszar1991least},
which is typically derived for Poisson noise.
The I-divergence based data term for the amplitude model is given by
\cite{SteidlT10, wensen14}
\begin{equation}\label{dataterm3}
D_3(u, f) = \langle \lambda (u^2 - 2f^2\text{log}u),1 \rangle \,,
\end{equation}
which is strictly convex w.r.t. $u$.

{Concerning the relation between \eqref{dataterm1} and \eqref{dataterm3}, by following the argument stated in
Steidl and Teuber's work \cite{SteidlT10}, we achieve similar equivalence by
incorporating these two data terms into a Total Variation (TV)
regularized variational model in the continuous form. The derivation reads as
follows. Regarding the data term \eqref{dataterm1}, let us consider the following convex
variational functional by setting $u = e^{\frac 1 2 w}$
\begin{equation}\label{SO-model}
\hat w = \arg\min\limits_{w \in BV} \int_\Omega \left(w + f^2 e^{-w}\right) dx +
\lambda |w|_{TV}, ~ \hat u = e^{\frac 1 2 \hat w}\,,
\end{equation}
where $|w|_{TV}$ is the TV semi-norm defined as
\begin{equation}\label{tv}
|w|_{TV} = \int_\Omega |\nabla w| dx\,.
\end{equation}

Regarding the data term \eqref{dataterm1}, let us consider the following convex
variational functional by setting $u = \sqrt{v}$
\begin{equation}\label{div-model}
\hat v = \arg\min\limits_{v \in BV, v > 0}
\int_\Omega \left(v -f^2\text{log}v \right) dx +
\lambda |v|_{TV}, ~ \hat u = \sqrt{v}\,.
\end{equation}

As $\nabla e^w = e^w \nabla w$, we have for $u = e^w$
that $\nabla u (x) = 0$ if and only
if $\nabla w (x) = 0$. As a consequence,
if we minimize \eqref{SO-model} and
\eqref{div-model} over smooth functions,
the minimizer $\hat w$ for \eqref{SO-model} and the minimizer $\hat v$ for
\eqref{div-model} are respectively given by
\[
1 - f^2e^{-\hat w} - \lambda \text{div}\frac{\nabla \hat w}{|\nabla \hat w|} = 0\,,
\] and
\[
1 - \frac{f^2}{\hat v} - \lambda \text{div}\frac{\nabla \hat v}{|\nabla \hat v|} = 0\,,
\]
for those points
$|\nabla \hat w (x)| \neq 0$ and $|\nabla \hat v (x)| \neq 0$.
Since we have $\frac{\nabla w}{|\nabla w|} = \frac{e^w \nabla w}{e^w |\nabla w|} =
\frac{\nabla u}{|\nabla u|}$, it turns out that
the minimizer $\hat w$ and $\hat v$ of the functional in the model \eqref{SO-model}
and model \eqref{div-model} respectively, coincide in the sense that
$\hat v = e^{\hat w}$. Therefore, for both $\hat u$, we will obtain the same result.
Note that this conclusion only holds for the those $L_1$-norm induced regularizers,
such as the TV regularization. For some other regularized variational models (e.g., TGV-based model \cite{wensen14}), although this data term \eqref{dataterm3} seems inappropriate,
it performs very well for despeckling.}

\begin{figure*}[t]
\centering
\vspace{-1cm}
{\includegraphics[width=0.8\linewidth]{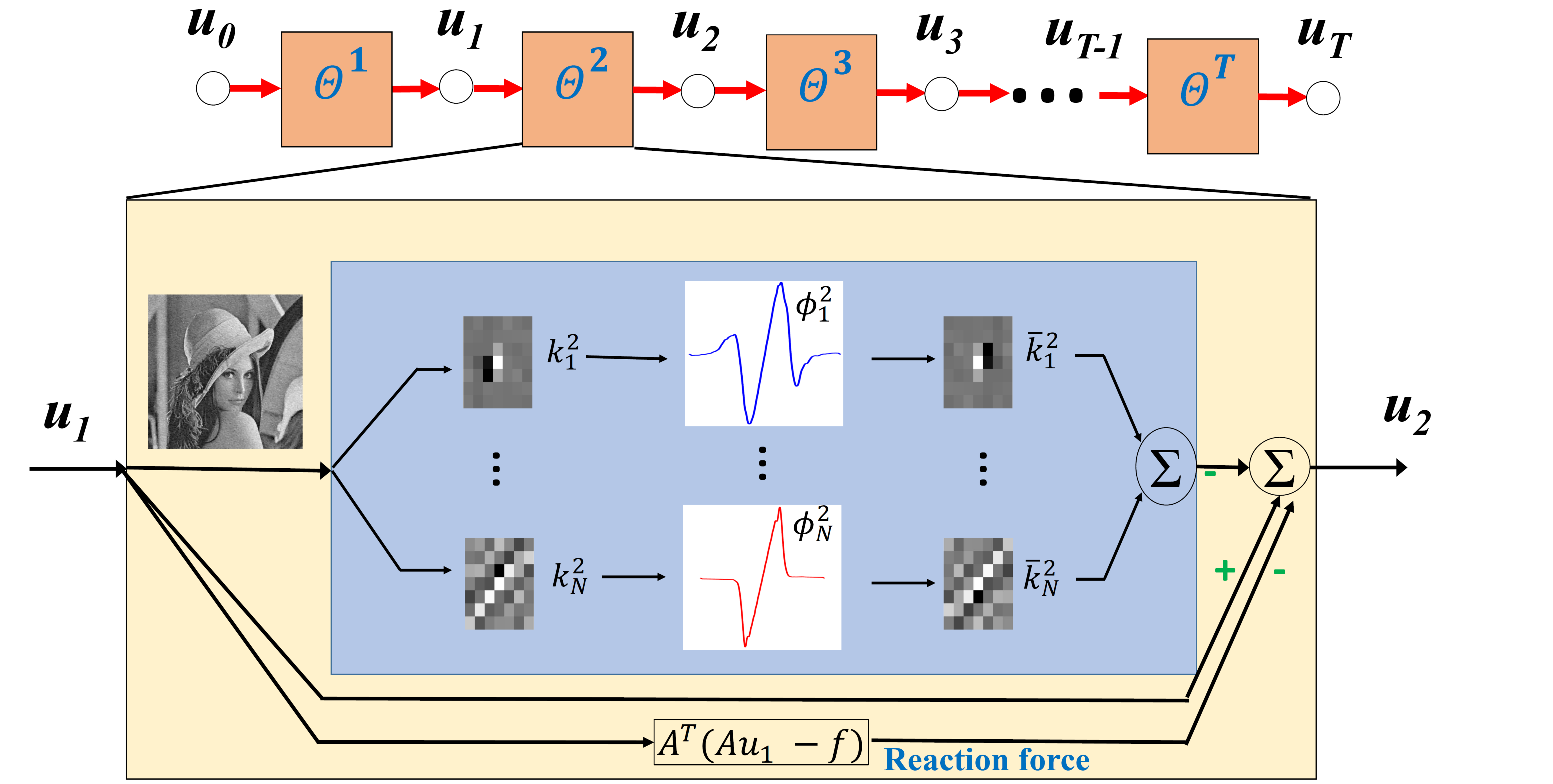}}
\vspace*{-0.4cm}
\caption{{The detailed architecture of the original TNRD model with a reaction term, e.g., $\psi(u_{t-1}, f) = A^\top (A u_{t-1} - f)$.
Note that it is represented as a feed-forward network.}}\label{fig:feedforwardCNN}
\end{figure*}

\subsection{Trainable Nonlinear Reaction Diffusion (TNRD)}
\subsubsection{Highly parametrized nonlinear diffusion model}
Recently, a simple but effective framework for image restoration called TNRD
was proposed in \cite{TNRD}, which is based on the concept of
nonlinear reaction diffusion.
The TNRD framework is modeled by highly parametrized
linear filters as well as highly parametrized
influence functions. In contrast to those
conventional nonlinear diffusion models which
usually make use of handcrafted parameters, all the
parameters in the TNRD model, including the filters and the influence functions, are learned from training data through a loss based approach.

The proposed framework is formulated as the following time-dynamic nonlinear
reaction-diffusion process with $T$ steps
\begin{equation}\label{diffusion}
\footnotesize
\begin{cases}
u_0 = I_0, \quad t = 1 \cdots T \\
u_{t} = \underbrace{\text{Prox}_{\cG^t}}_\text{reaction force}\left(
u_{t-1} - \left(\underbrace{\sum\limits_{i = 1}^{N_k}\bar k_i^t * \phi_i^t
(k_i^t * u_{t-1})}_\text{diffusion force} + \underbrace{\psi^t(u_{t-1}, f)}_
\text{reaction force}\right)\right)\,,
\end{cases}
\end{equation}
where $I_0$ is the initial status of the diffusion process, $*$ is the convolution operation, $T$ denotes the diffusion stages,
$N_k$ denotes the number of filters, $k_i^t$ are time varying
convolution kernels, $\bar{k}_i$ is obtained by rotating the kernel
$k_i$ 180 degrees, $k_i*u$ denotes 2D convolution of the image $u$
with the filter kernel $k_i$, and $\phi_i^t$ are time varying
influence functions (not restricted to be of a certain kind).
Both the proximal mapping operation $\text{Prox}_{\cG^t}(\hat u)$
and $\psi^t(u_t, f)$ are the reaction force.
Usually, the reaction term $\psi^t(u)$ is chosen as
the derivative of a certain
smooth date term $\cD(u, f)$, i.e., $\psi^t(u) = \nabla_u \cD^t(u)$. Note that
the proximal mapping operation \cite{nesterov2004introductory} related to
the function $\cG^t$ is given as
\[
\text{Prox}_{\cG^t}(\hat u) = \min\limits_{u}\frac{\|u - \hat u\|_2^2}{2}
+ \cG^t(u, f) \,.
\]

As shown in \cite{TNRD},
the proposed model \eqref{diffusion} can be interpreted as
performing one gradient descent step at $u^t$ with respect to a dynamic energy
functional given by
\begin{equation}\label{foemodel}
E^t(u, f) = \suml{i = 1}{N_k}\sum\limits_{p = 1}^{N} \rho_i^t((k_i^t
*u)_p) + \cD^t(u, f) + \cG^t(u, f)\,,
\end{equation}
where the functions $\left\{\rho_i^t\right\}_{t=0}^{t=T-1}$ are the so-called penalty
functions and the regularization parameter is included in $\cD^t(u, f)$. Note that $\rho'(z) = \phi(z)$ and the parameters \{$k_i^t,
\rho_i^t$\} vary across the stages i.e., changes at each iteration.

It is easy to apply the proposed TRND framework to
different image restoration problems
by incorporating specific reaction force, such as
Gaussian denoising, image deblurring, image super resolution and image
inpainting. This is realized by setting $\cG^t = 0$, while $\cD^t(u, f) =
\frac{\lambda^t}{2} \|A u - f\|_2^2$ and $\psi^t(u) = \lambda^t A^\top(Au -
f)$,
where $\lambda^t$ is related to the strength of the reaction term, $u$ and $f$ denote the original true image and the input degraded image, respectively, and $A$ is the
associated different linear operator. In the case of Gaussian denoising, $A$ is the
identity matrix; for image super resolution, $A$ is related to the down sampling operation and for image
deconvolution, $A$ should correspond to the linear blur kernel.

\subsubsection{Overall training scheme}
The proposed diffusion model in \cite{TNRD} is trained in a supervised manner. In other words, the input/output pairs
for certain image processing task are firstly prepared, and then
we exploit a loss minimization scheme to
optimize the model parameters $\Theta^t$ for each
stage $t$ of the diffusion process. The training dataset consists of
$S$ training samples $\{u_{gt}^s,f^s\}_{s=1}^S$, where
$u_{gt}^s$ is a ground truth image and $f^s$ is the corresponding degraded input.
The model parameters in stage $t$ required
to be trained include 1) the reaction force weight $\lambda$,
(2) linear filters and (3) influence functions. All parameters are grouped
as $\Theta^t$, i.e., $\Theta^t = \{\lambda^t, \phi_i^t, k_i^t\}$.
Then, the optimization problem for the training task is formulated as follows
\begin{equation}\label{learning}
\small
\hspace{-0.25cm}
\begin{cases}
\min\limits_{\Theta}\cL(\Theta) = \sum\limits_{s = 1}^{S} \ell(u_T^s, u_{gt}^s) =
\sum\limits_{s = 1}^{S}\frac 1 2
\|u_T^s - u_{gt}^s\|^2_2\\
\text{s.t.}
\begin{cases}
u_0^s = I_0^s \\
u_{t}^s = \text{Prox}_{\cG^t}\left(u_{t-1}^s - \left({\sum\limits_{i = 1}^{N_k}
\bar k_i^t * \phi_i^t(k_i^t * u_{t-1}^s)} + \psi^t(u_{t-1}^s, f^s)
\right)\right), \\
t = 1 \cdots T\,,
\end{cases}
\end{cases}
\end{equation}
where $\Theta = \{\Theta^t\}_{t=1}^{t=T}$.
The training problem in \eqref{learning} can be solved via gradient
based algorithms, e.g., the L-BFGS algorithm \cite{lbfgs}, where the gradients associated with $\Theta_t$ are computed
using the standard back-propagation technique \cite{lecun1998gradient}.

There are two training strategies to learn the diffusion processes: 1) the greedy training strategy to learn the diffusion process stage-by-stage;
 and 2) the joint training strategy to train a diffusion process by simultaneously tuning
the parameters in all stages. Generally speaking, the joint training strategy performs better \cite{TNRD}, and the greedy training strategy is often
 used to provide a good initialization for the joint training.
Concerning the joint training scheme, the gradients $\frac {\partial
\ell(u_T, u_{gt})}{\partial \Theta_t}$ are computed as follows,
\begin{equation}
\frac {\partial \ell(u_T, u_{gt})}{\partial \Theta_t} =
\frac {\partial u_t}{\partial \Theta_t} \cdot \frac {\partial u_{t+1}}{\partial u_{t}} \cdots
\frac {\partial \ell(u_T, u_{gt})}{\partial u_T} \,.
\label{iterstep}
\end{equation}
For different image restoration problems, we mainly need to
recompute the two components $\frac {\partial u_t}{\partial \Theta_t}$ and
$\frac {\partial u_{t+1}}{\partial u_{t}}$, the main part of which
are similar to the derivations in \cite{TNRD}.
\section{Optimized Nonlinear Diffusion Filtering for Despeckling}
\label{section3}
\subsection{Proposed Trainable Reaction Diffusion Filtering for Despeckling}
\label{despeckling_model}
{
The diffusion filtering process for despeckling can be derived from
the energy functional \eqref{foemodel} by incorporating data fidelity terms in
\eqref{dataterm1}-\eqref{dataterm3}.

Considering the data term \eqref{dataterm2},
we have the following energy functional
\begin{equation}
E(w, f) = \suml{i = 1}{N_k}\sum\limits_{p = 1}^{N} \rho_i((k_i
*e^w)_p) + \lambda \langle 2w + f^2 e^{-2w},1 \rangle\,,
\label{foedataterm2}
\end{equation}
with $u = e^w$. Then, we derive the corresponding diffusion process by setting
$\cD(w, f) = \lambda \langle 2w + f^2 e^{-2w},1 \rangle$ and $\cG = 0$, as
the proximal mapping operation w.r.t $\cG (w,f) = \lambda \langle 2w + f^2 e^{-2w},1 \rangle$
is not easy to compute. The resulting diffusion process is given as
\[
w_{t} = w_{t-1} - \left(
\begin{array}{l}
\sum\limits_{i = 1}^{N_k}
e^{w_{t-1}} \odot \left(\bar k_i^t * \phi_i^t(k_i^t * e^{w_{t-1}})\right) +
{\lambda^t}(2 - 2f^2e^{-2w_{t-1}})
\end{array}
\right) \,,
\]
where $\odot$ denotes element-wise product. However, due to the formulation
$e^{w_{t-1}} \odot \left(\bar k_i^t * \phi_i^t(k_i^t * e^{w_{t-1}})\right)$,
the value of $e^{w_T}$ is prone to explosion in training phase and therefore makes
the training fail.

If we consider the data term \eqref{dataterm1}, we arrive at the following energy functional
\begin{equation}\label{withdata1}
E(u, f)|_{u > 0} = \suml{i = 1}{N_k}\sum\limits_{p = 1}^{N} \rho_i((k_i
*u)_p) + \lambda
\langle 2\text{log} u + \frac {f^2}{u^2},1 \rangle\,.
\end{equation}
As the proximal mapping operator $\text{Prox}_{\cG}(\cdot)$ only works for a
convex function $\cG$, we have to set
$\cD(u, f) = \lambda \langle 2\text{log} u + \frac {f^2}{u^2},1 \rangle$ and
$\cG = 0$ in \eqref{foemodel}. Then we arrive at a direct gradient descent process
with ${\psi(u_t, f)} = \lambda\left( \frac 2 u - \frac{2f^2}{u^3}\right)$.
However, this straightforward gradient descent approach
is not applicable in practice, because (1) at the points
with $u$ very close to zero the reaction term is enlarged so much that
there will be an obvious problem of numerical instability;
(2) this update rule may produce negative values of
$u$ after one diffusion step, which will violate the constraint of the
data term in \eqref{withdata1}.

\begin{figure*}[t]
\centering
\vspace*{-3cm}
\hspace*{-0.9cm} {\includegraphics[width=0.8\linewidth]{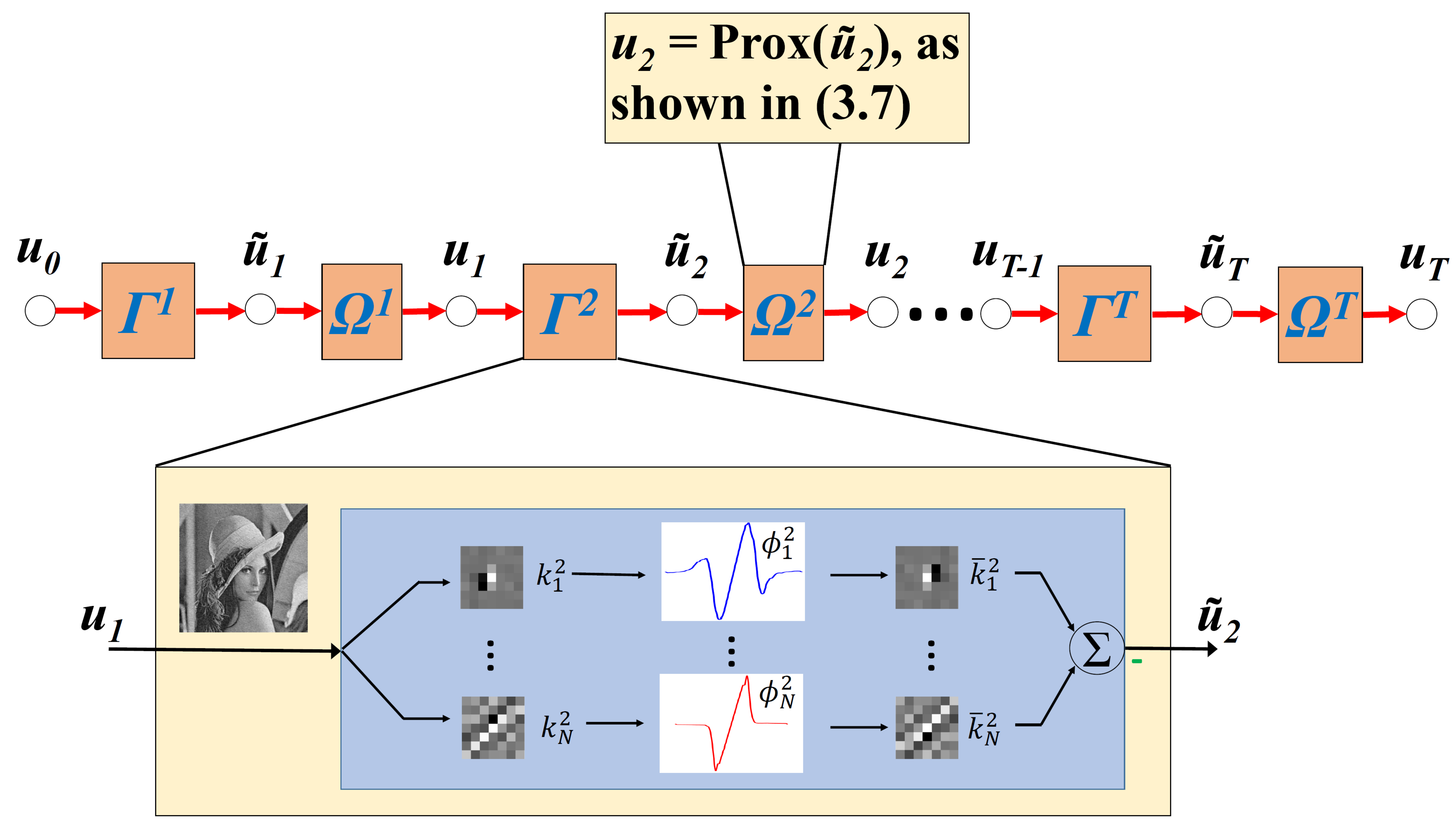}}
\vspace*{-0.3cm}
\caption{{The architecture of the proposed diffusion model for despeckling. It is represented as a feed-forward network. Note that the additional convolution step with the rotated kernels $\bar k_i$. Here, the training parameters for the $t$th step are $\mathit \Theta^t = \left\{\mathit \Gamma^t, \mathit \Omega^t \right\}$ with $\mathit \Gamma^t = \{ \phi_i^t, k_i^t \}_{i=1}^{N_k}$ and $\mathit \Omega^t = \lambda^t$ respectively. More importantly, the employed scheme based on the proximal gradient descent method is quite different from the straightforward direct gradient descent (as shown in Fig.~\ref{fig:feedforwardCNN}) employed for TNRD-based denoisng task \cite{chenCVPR15}}.}
\label{fig:feedforwardCNNspeckle}
\end{figure*}

\begin{figure}[t]
\centering
\hspace*{-0.9cm} {\includegraphics[width=0.5\linewidth]{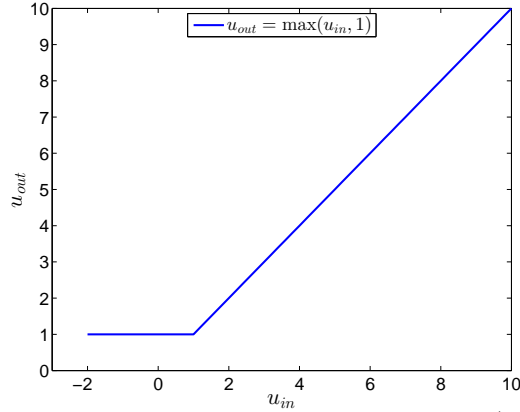}}
\vspace*{-0.5cm}
\caption{{The hard shrinkage operation $u_{out} = \mathrm{max}(u_{in},c)$ with $c=1$.}}
\label{maxfun}
\end{figure}

{In order to remedy the aforementioned problems, after each diffusion step, we additionally consider
a simple projection operation
\[
u= \max(\tilde{u}, c),~\text{with}~ c > 0\,,
\]
to shrink those ``bad'' pixels, which do not satisfy the constraint $u > 0$ or are
very close to zero. Therefore, we arrive at the following diffusion approach
\begin{equation}\label{diffusion_backP}
\begin{split}
\tilde{u}_{t+1} & = u_t - \left(\suml{i=1}{N_k}\bar{k}_i^{t+1} *
\phi_i^{t+1}(k_i^{t+1} * u_{t}) +
\frac {\lambda^{t+1}}{u_t}\left( 1 - \frac{f^2}{u_t^2}\right)\right) \,,\\
{u}_{t+1} & = \max(\tilde{u}_{t+1}, c)\,.
\end{split}
\end{equation}
An illustrative example of the shrinkage function related to the projection
operation is show in Fig.~\ref{maxfun}, where we set $c = 1$.
Note that as the shrinkage function is
differentiable, it is not problematic in the training phase.

As the projection operation will somehow manipulate the pixel values, a small
$c$ is required to reduce its influence. However, on the other hand, a small $c$
will raise a problem of numerical stability due to the formulation
$\frac 2 u - \frac{2f^2}{u^3}$, and then training will fail.
In order to search an appropriate $c$ for our diffusion model, we progressively
increased $c$ from $0.1$ with four settings $c = 0.1, 0.2, 0.5, \text{and}~ 1$.
We observed that the former three settings led to failure in the training phase, and
$c = 1$ is sufficient for a stable training.

However, although the diffusion model \eqref{diffusion_backP} overcomes the shortcomings of the model \eqref{dataterm1}, it encounters some other problems which will be explained in detail in the experimental part \ref{compareexperi}.}

As a consequence, we resort to the data term \eqref{dataterm3}.
Although the data term \eqref{dataterm3} sounds inappropriate,
it is incorporated in a flexible training framework with many free parameters.
The training phase can somehow automatically adapt to this inappropriate data term.

Therefore, we exploit the data term \eqref{dataterm3}, and reach the following
energy functional
\begin{equation}\label{straightforward}
\min\limits_{u > 0}E(u) =
\suml{i=1}{N_k} \sum\limits_{p = 1}^{N} \rho_i((k_i
*u)_p) + \lambda \langle u^2 - 2f^2\text{log}u,1 \rangle \,.
\end{equation}

As the data term in \eqref{straightforward} is smooth and differentiable
we have two possible ways to derive the diffusion process.
\begin{itemize}
\item[I)] By setting $\cD(u, f) = \lambda \scal{u^2-2f^2\text{log}u}{1}$ and
$\cG = 0$ in \eqref{foemodel}, we arrive at a direct gradient descent process
with ${\psi(u_t, f)} = \lambda\left( 2u-\frac{2f^2}{u}\right)$. In this case,
we will encounter the same problem mentioned above, thus intractable in practice.
\item[II)] By setting $\cD = 0$ and
$\cG(u, f) = \lambda \scal{u^2-2f^2\text{log}u}{1}$ in \eqref{foemodel}, the resulting diffusion process
is able to overcome the above shortcomings. Details read as follows.
\end{itemize}
}

Recall that the proximal mapping with respect to $\cG$ is given as
the following minimization problem
\begin{equation}\label{subproblemGpoisson}
\left( \mat{I} +  \partial \cG \right)^{-1}(\tilde{u}) =
\arg\min\limits_{u} \frac{\|u - \tilde
{u}\|^2_2}{2} +  \lambda \langle u^2-2f^2\mathrm{log}u,1 \rangle \,.
\end{equation}
The solution of \eqref{subproblemGpoisson} is given by the following point-wise operation
\begin{equation}\label{subproblemGIdiv}
\hat u = \left( \mat{I} +  \partial \cG \right)^{-1}(\tilde{u})  = \frac{ \tilde{u}
+\sqrt{\tilde{u}^2 + 8\left( 1+2 \lambda \right)  \lambda f^2}}{2\left( 1+2 \lambda \right)} \,.
\end{equation}
Note that this update rule is able to
guarantee $\hat u > 0$ in diffusion steps because $\hat u$ is always positive if $f > 0$.

Finally, the diffusion process for despeckling using
the proximal gradient method is formulated as
\begin{equation}\label{diffusionprocessfinal}
u_{t+1}  = \frac{ \tilde{u}_{t+1}
+\sqrt{\tilde{u}_{t+1}^2 + 8\left( 1+2\lambda^{t+1} \right) \lambda^{t+1} f^2}}{2\left( 1+2 \lambda^{t+1} \right)} \,,
\end{equation}
where $\tilde{u}_{t+1} =
u_t - \suml{i=1}{N_k}\bar{k}_i^{t+1} * \phi_i^{t+1}(k_i^{t+1} * u_{t})$. The architecture of the proposed diffusion model for despeckling is as shown in Fig.~\ref{fig:feedforwardCNN}.

\subsection{Computing The Gradients for Training}
In this section we derive the gradients of the
loss function w.r.t the training parameters $\Theta_t = \left\{ \lambda^t,\phi_i^t,
k_i^t \right\}$ in joint training.
In summary, we need to compute three parts
of $\frac {\partial \ell(u_T, u_{gt})}{\partial \Theta_t}$,
i.e., $\frac {\partial u_{t+1}}{\partial u_{t}}$, $\frac {\partial u_t}{\partial \Theta_t}$ and $\frac {\partial \ell(u_T, u_{gt})}{\partial u_T}$.

First of all, according to \eqref{learning}, it is easy to check that
the gradient $\frac {\partial \ell(u_T, u_{gt})}{\partial u_T}$ is
given as follows
\[
\frac {\partial \ell(u_T, u_{gt})}{\partial u_T} = u_T - u_{gt}\,,
\]
where we omit the image index $s$ for brevity. Then,
$\frac{\partial u_{t+1}}{\partial u_t}$ is computed as follows according to the chain rule,
\begin{equation}\label{derv1}
\frac{\partial u_{t+1}}{\partial u_t} =\frac{\partial \tilde{u}_{t+1}}{\partial u_t} \cdot \frac{\partial u_{t+1}}{\partial \tilde{u}_{t+1}}.
\end{equation}
According to \eqref{diffusionprocessfinal},
it is easy to check that
\begin{equation}\label{derv1-1}
\begin{array}{l}
\frac{\partial \tilde{u}_{t+1}}{\partial u_t} = \mat{I} -
\suml{i=1}{N_k} {K_i^{t+1}}^\top \cdot \Lambda_i \cdot {\left({\bar{K}_i}^{t+1}\right)}^\top.
\end{array}
\end{equation}
where $\Lambda_i$ is a diagonal matrix
$\Lambda_i = \text{diag}({\phi_i^t}'(z_1), \cdots, {\phi_i^t}'(z_p))$ with ${\phi_i^t}'$ denoting the first order derivative of function ${\phi_i^t}$ and
$z = {k_i^{t+1}} * u_t$. Here, $\{ z_i \}_{i=1}^{i=p}$ denote the elements of $z$ which are represented as a column-stacked vector.
Note that ${\bar{K}_i}$ is related to the kernel
$\bar{k}_i$, i.e., $\bar{K}_i u \Leftrightarrow \bar k_i*u$. As shown in \cite{TNRD}, $K_i^\top$ and
${\bar{K}_i}^\top$ can be computed by the convolution operation
with the kernel $k_i$ and $\bar{k}_i$, respectively with careful boundary handling. Moreover, $\frac{\partial u_{t+1}}{\partial \tilde{u}_{t+1}}$ can be easily derived according to \eqref{diffusionprocessfinal},
and is formulated as
\begin{equation}\label{derv2}
\frac{\partial u_{t+1}}{\partial \tilde{u}_{t+1}} =
\text{diag}(y_1, \cdots, y_p)\,,
\end{equation}
where $\{ y_i \}_{i=1}^{i=p}$ denote the elements of
\[
y = \frac{1}{2\left( 1+2 \lambda^{t+1} \right)} \left[ 1+\frac{\tilde{u}_{t+1}}
{\sqrt{\tilde{u}_{t+1}^2 + 8\left( 1+2\lambda^{t+1} \right) \lambda^{t+1} f^2}} \right]\,.
\]
Now, the calculation of $\frac{\partial u_{t+1}}{\partial u_t}$
is obtained by incorporating \eqref{derv1-1} and \eqref{derv2}.

Concerning the gradients $\frac{\partial u_t}{\partial \Theta_t}$, we should derive three parts of $\Theta_t=\left\{ \lambda^t,\phi_i^t, k_i^t \right\}$
respectively. It is worthy noting that the gradients of $u_t$ w.r.t $\left\{\phi_i^t, k_i^t \right\}$ are only associated with $\tilde{u}_t=u_{t-1} - \suml{i=1}{N_k}\bar{k}_i^t * \phi_i^t(k_i^t * u_{t-1})$, and are given as
$\frac{\partial u_t}{\partial \phi_i^t} = \frac{\partial \tilde{u}_t}{\partial \phi_i^t} \cdot  \frac{\partial u_t}{\partial \tilde{u}_t},$
and
$\frac{\partial u_t}{\partial k_i^t} =  \frac{\partial \tilde{u}_t}{\partial k_i^t} \cdot \frac{\partial u_t}{\partial \tilde{u}_t}.
$

The detailed derivations of $\frac{\partial \tilde{u}_t}{\partial \phi_i^t}$
has been provided in our previous work \cite{TNRD}, and therefore we
omit them in this paper.
The formulation of $\frac{\partial u_t}{\partial \tilde{u}_t}$ is
straightforward to compute according to \eqref{derv2}.

According to \eqref{diffusionprocessfinal}, the gradient of $u_t$ w.r.t $\lambda^t$ is formulated as
\begin{equation}\label{derv4}
\frac{\partial u_t}{\partial \lambda^t} =
(x_1, \cdots, x_p)\,,
\end{equation}
where $\{ x_i \}_{i=1}^{i=p}$ denote the elements of
\begin{equation}
\footnotesize
x=
\frac{2f^2+8\lambda^{t}f^2}{\left(1+2\lambda^{t}\right)\sqrt{\tilde{u}_{t}^2 + 8\left( 1+2\lambda^{t} \right) \lambda^{t} f^2}}
-
\frac{\tilde{u}_{t} + \sqrt{\tilde{u}_{t}^2 + 8\left( 1+2\lambda^{t} \right) \lambda^{t} f^2}}{ \left( 1+2\lambda^{t} \right)^2 }
\,.
\end{equation}
Note that $\frac{\partial u_t}{\partial \lambda^t}$ is written as
a row vector.

In practice, in order to ensure the value of $\lambda^t$ positive during
the training phase, we set $\lambda=e^{\beta}$.
As a consequence, in implementation we employ the
gradient $\frac{\partial u_t}{\partial \beta^t}$ instead of
$\frac{\partial u_t}{\partial \lambda^t}$. The gradient $\frac{\partial u_t}
{\partial \beta^t}$ is explicitly formulated as
\begin{equation}\label{derv5}
\small
\frac{\partial u_t}{\partial \beta^t} = \lambda^t \left[
\begin{array}{l}
\frac{2f^2+8\lambda^{t}f^2}{\left(1+2\lambda^{t}\right)\sqrt{\tilde{u}_{t}^2 + 8\left( 1+2\lambda^{t} \right) \lambda^{t} f^2}}
-
\frac{\tilde{u}_{t} + \sqrt{\tilde{u}_{t}^2 + 8\left( 1+2\lambda^{t} \right) \lambda^{t} f^2}}{ \left( 1+2\lambda^{t} \right)^2 }
\end{array}
\right].
\end{equation}
{The architecture of the proposed diffusion model for despeckling is as shown in Fig.~\ref{fig:feedforwardCNNspeckle} which is quite different from the straightforward direct gradient descent (as shown in Fig.~\ref{fig:feedforwardCNN}) used for the original TNRD-based denoisng task \cite{TNRD}.}

\section{Experiments}
\label{experiments}
In this section, we abbreviate the proposed method as TRDMD (the fully Trained Reaction Diffusion Models for Despeckling). To evaluate the performance of the proposed
method, three representative state-of-the-art approaches are compared: the nonlocal methods PPBit \cite{deledalle2009iterative}, SAR-BM3D \cite{ParrilliPAV12}, and our previous FoE-based despeckling work \cite{chen2014higher}.
Moreover, considering that the proposed model is essentially a diffusion process, we also employ the traditional $\Gamma$-Map filter \cite{lopes1990maximum} and some related anisotropic diffusion approaches for comparison, e.g., speckle
reducing anisotropic diffusion (SRAD) \cite{yu2002speckle} and detail preserving anisotropic diffusion
(DPAD) \cite{aja2006estimation}. The corresponding codes are downloaded from the authors' homepage,
and we use them as is. {Note that the two excellent variational methods \cite{zengtie1} and \cite{zengtie2} mainly focus on the restoration
of images that are simultaneously blurred and corrupted by multiplicative noise. Even though these two papers also work for the problem of pure multiplicative noise
reduction, the more advanced approaches such as patch-based method PPBit or
SAR-BM3D for multiplicative noise removal could have better
results. In \cite{zengtie1},
a TV regularized variational model is exploited, which is a pair-wise model and
inevitably suffers from the well-known stair-case effect. In our model,
we exploited a much more effective image regularization term with larger clique
and adjustable potential functions. In \cite{zengtie2}, the nonlocal filtering algorithm proposed in \cite{teuber2012new} is
exploited to accomplish the first step for removing multiplicative noise.
The nonlocal filtering algorithm \cite{teuber2012new} makes use of the same
framework as PPBit, but investigates a different similarity measure in the
presence of multiplicative noise. It is stated in \cite{deledalle2012compare} that
the nonlocal filtering algorithms \cite{teuber2012new} and PPBit
both have the same performance. As a consequence, we only provide comparison to those algorithms PPBit, SAR-BM3D and FoE-based methods, which generally perform better than \cite{zengtie1} and \cite{zengtie2}.}

These methods are evaluated on several real SAR images and synthesized images with speckle noise using 68 optical images originally introduced by \cite{RothFOE2009}, which have been widely exploited for Gaussian denoising. To provide a comprehensive comparison, the test number of looks are distributed between 1 to 8. For the quality assessment of
despeckling methods, we closely follow the indexes employed in \cite{argenti2013tutorial}. Two classes of indexes are involved: full-reference quality indexes for simulated SAR images and no-reference quality indexes for real SAR images. Three commonly used with-reference indexes are taken to evaluation, \ie, PSNR, the mean structural similarity index (MSSIM) \cite{ssim} and the edge correlation (EC) \cite{sattar1997image} \cite{achim2003sar}. The MSSIM underlines the perceived changes in structural information after despeckling. MSSIM takes values over the interval [0,1], where 0 and 1 indicate no structural similarity and perfect similarity, respectively. The EC index is a measure of edge preservation between the high-pass versions of the original and filtered images. Larger values
correspond to a better edge retaining ability of the despeckling method.

For without-reference indexes, we employ the mean $RI_M$ and the variance $RI_V$ of the ratio image $RI$ \cite{oliver2004understanding}. The ideal values of $RI_M$ and $RI_V$ are one and $(4/\pi-1)/L$ respectively for an $L-$look amplitude SAR image \cite{oliver2004understanding}. We also employ the comparison between the coefficient of variation calculated on the despeckled image, namely $C_{\hat u}$, and its expected theoretical value on the noise-free image, $C_u$ \cite{touzi2002review}. The coefficient of variation is a widespread indicator of texture preservation. Theoretically speaking, the value of $C_{\hat{u}}$ should be close to $C_u$. If $C_{\hat{u}}$
departs significantly from $C_u$ the texture is certainly altered.

In the following, the nonlinear diffusion process of stage $T$
with filters of size $m\times m$ is expressed as $\mathrm{TRDMD}_{m \times m}^T$ whose number of filters is $m^2-1$ in each stage, if not specified.

To generate the training data for the simulation denoising experiments, we cropped a 256$\times$256 pixel region
from each image of the Berkeley segmentation
dataset \cite{MartinFTM01}, resulting in a total of 400 training samples of size 256 $\times$ 256.
We also employ different amounts of training samples to observe the denoising performance comparison, as shown in
Fig.~\ref{influence}(a). Note that the PSNR values in the following three subsections are evaluated by averaging denoised results of 68 test images.
For simplicity the value of $L$ is set as 8 in the following three contrastive analysises, without loss of generality.
\begin{figure}[t!]
\centering
{\includegraphics[width=0.5\linewidth]{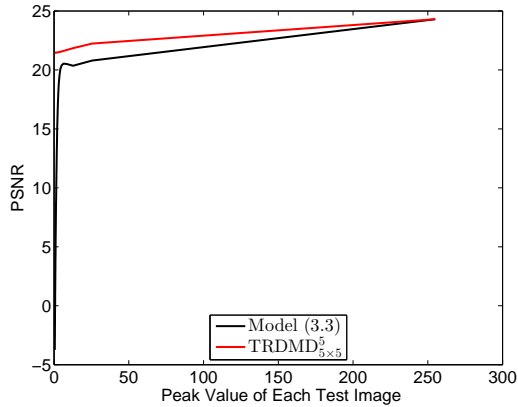}}
\caption{{Performance comparison between the proposed diffusion model $\mathrm{TRDMD}_{5 \times 5}^{5}$ and Model \eqref{diffusion_backP} for despeckling. The peak values of the test images are distributed between 0.5$\sim$255.}}
\label{compare}
\end{figure}
\subsection{The Experimental Comparison Between the Diffusion Model \eqref{diffusion_backP} and $\mathrm{TRDMD}$}
\label{compareexperi}
{In this subsection, we conducted experiments based on the diffusion model \eqref{diffusion_backP},
and compared its performance with the proposed model $\mathrm{TRDMD}$ in this study.
For the sake of fairness, these methods are evaluated on synthesized
images with speckle noise using 68 optical images, just as in the submitted paper.
The test number of looks $L$ is set as 1. In order to perform a
fast comparison, we employ the filter size $5 \times 5$ and stage = 5 to setup the
training models. Two cases as follows are considered for experimental analysis.

Case 1:

In this case, we take experiments directly on the 68 optical images whose minimum pixel value for each image is 1. The PSNR value obtained by the Model \eqref{diffusion_backP} is 24.30dB, just the same as $\mathrm{TRDMD}_{5 \times 5}^{5}$, indicating that the Model \eqref{diffusion_backP} also works for images with pixel values bigger than 1.

Case 2:

In this case, we set the maximum value (peak value) of each of the 68 optical images to be 0.5$\sim$255 by dividing the ground-truth image $u_{gt}$ by a factor $n$, i.e., $u_{gt}=\frac{u_{gt}}{n}$. The obtained PSNR values along with the peak values are presented in Figure~\ref{compare}. From Figure~\ref{compare} we can see that the gap between the two models decreases as the peak value increases. For lower peak values, the number of pixels that are smaller than 1 is bigger, leading to worse PSNR performance for Model \eqref{diffusion_backP}. Especially, there is an inflection point around peak value 6. At this point, the number of pixels that are smaller than 1 becomes so large that the performance of Model \eqref{diffusion_backP} is degraded obviously.

In summary, in order to perform a stable training, we have to set $c = 1$. However,
the hard projection operation $u = \mathrm{max}(u,1)$ will
obviously manipulate some of the pixels, because it will set those pixels that
should be 0 or very close to 0 to be 1.
If the number of those pixels is relatively small, it is not a problem.
However, if there is a large number of dark areas in the target image,
e.g., for the case of real SAR images, the projection $u = \mathrm{max}(u,1)$
will significant affect the despeckled results. As a consequence,
the despeckling performance will degrade dramatically.
In an extreme case, if we carry out a toy test on a constant image whose size
is $256 \times 256$ and pixel value is 0.5. The PSNR value obtained by the Model \eqref{diffusion_backP} is -0.01dB. However, the PSNR value obtained by
$\mathrm{TRDMD}_{5 \times 5}^{5}$ is 20.04dB, which is much higher than the
Model \eqref{diffusion_backP}.

Therefore, we mainly exploited the model $\mathrm{TRDMD}$ in this study.}

\subsection{Influence of The Number of Training Samples}
In this subsection, we evaluate the despeckling performance of trained models using different amounts of training samples for $\mathrm{TRDMD}_{5 \times 5}^5$.

\begin{figure}[t]
\centering
\subfigure[]{
\centering
\includegraphics[width=0.31\textwidth]{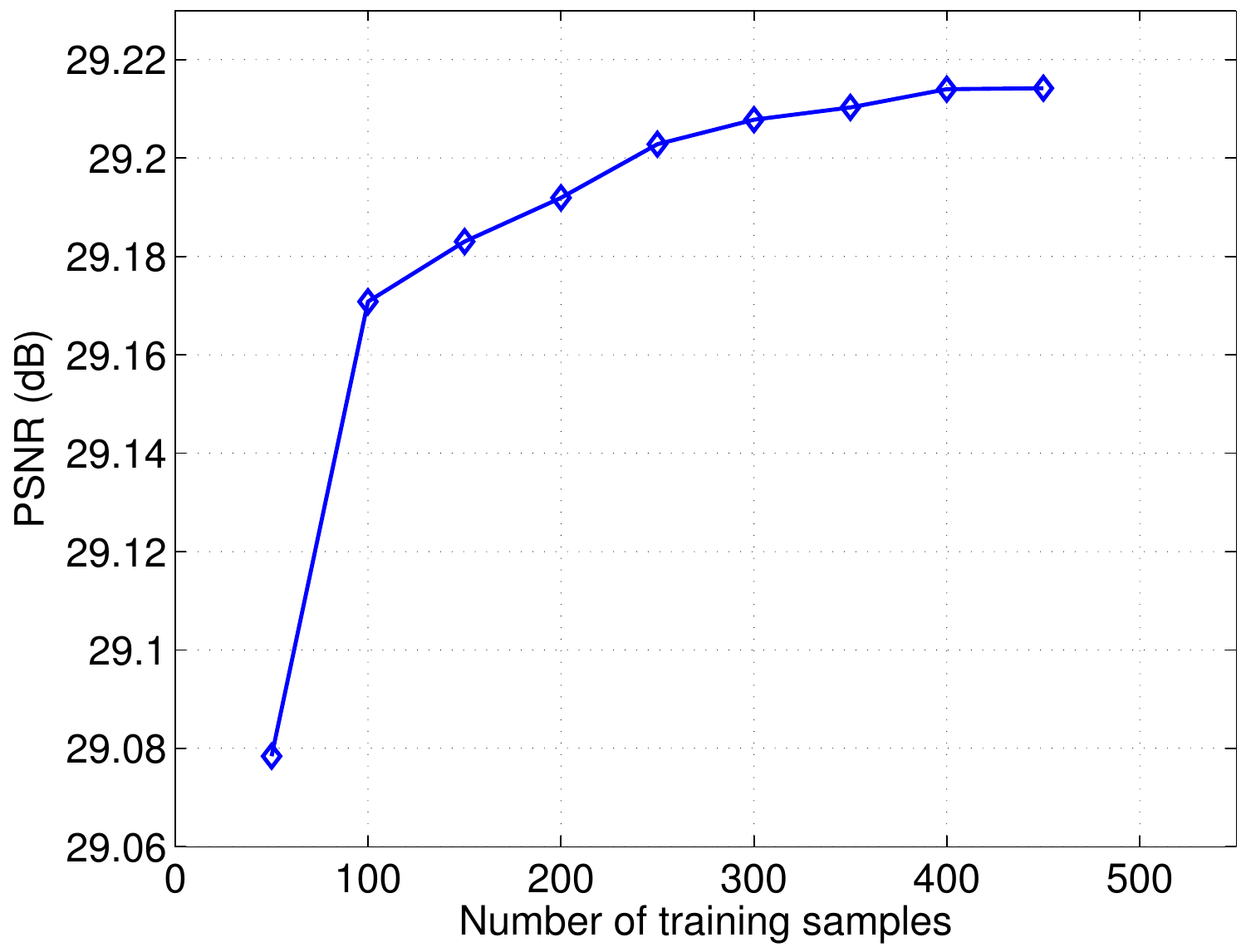}}
\subfigure[]{
\centering
\includegraphics[width=0.31\textwidth]{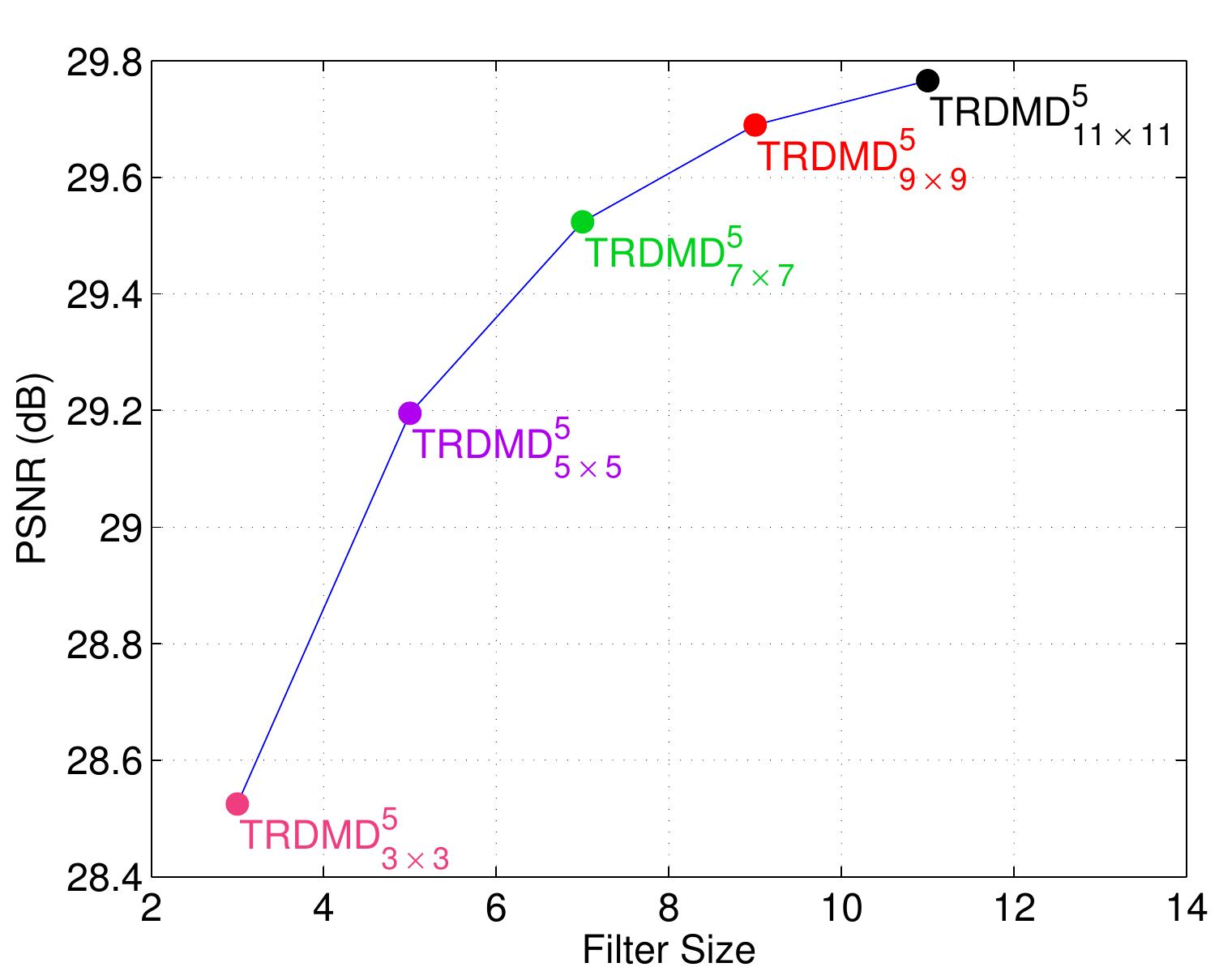}}
\subfigure[]{
\centering
\includegraphics[width=0.31\textwidth]{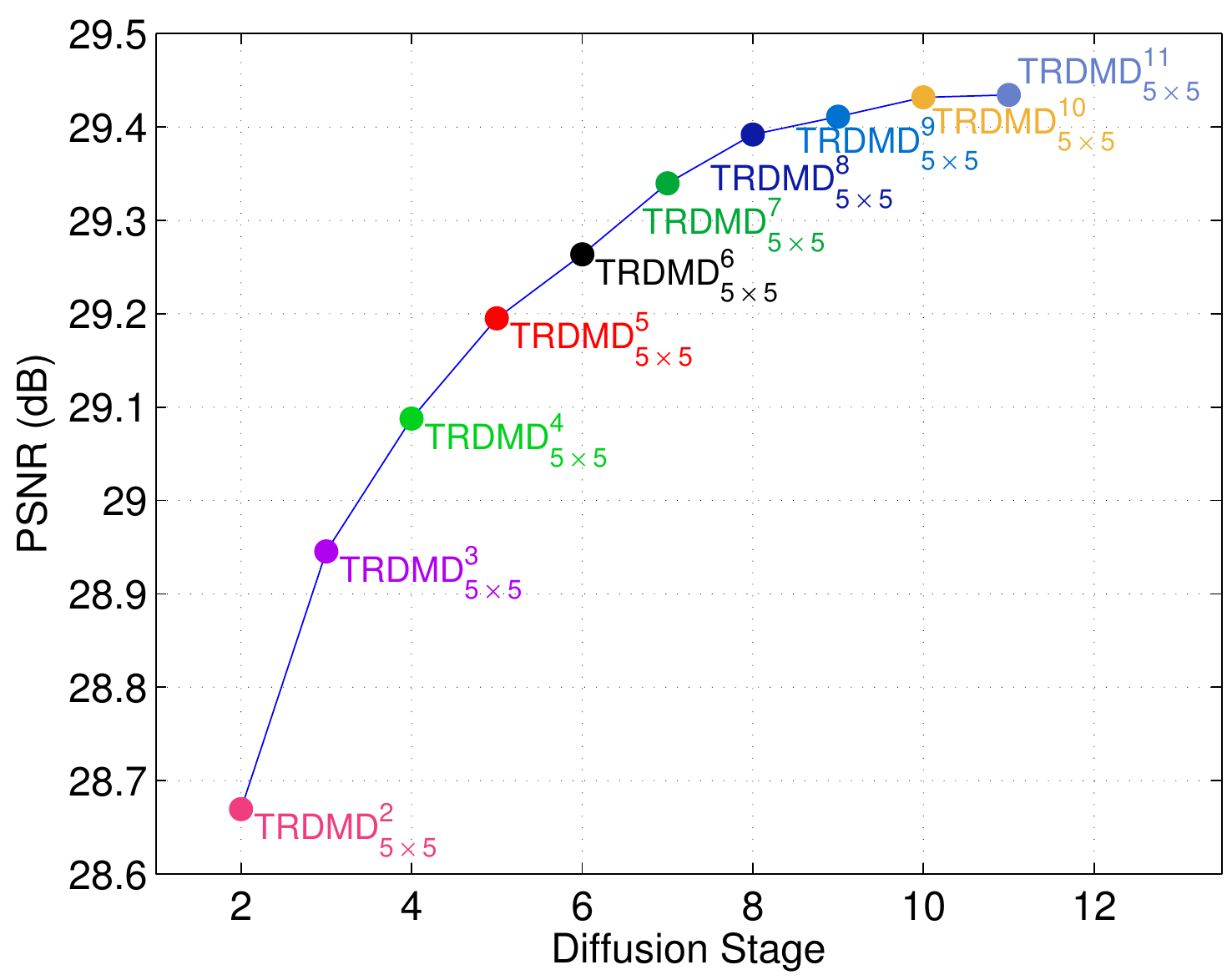}}
\caption{(a) Influence of the number of training examples. (b) Influence of the filter size. (c) Influence of the number of diffusion stages.}
\label{influence}
\end{figure}

The results are summarized in Fig.~\ref{influence}(a), from which one can see that too small training set will result in over-fitting which leads to inferior PSNR value.
By observation, 400 images are typically enough to provide reliable performance.

\subsection{Influence of filter size}
In this subsection, we investigate the influence of the filter size on the despeckling performance in Fig.~\ref{influence}(b).
The diffusion stages are set as 5, and 400 images are used for training.

One can see that increasing
the filter size keeps bringing improvement. However, the rate of increase in PSNR becomes relatively slower for larger filter size.
Balancing the training time and the performance improvement, we choose the $\mathrm{TRDMD}_{7 \times 7}^T$ model in our experiments.

\begin{figure*}[htbp]
\centering
\subfigure[{\scriptsize{$Mountain$}}]{
\centering
\includegraphics[width=0.3\textwidth]{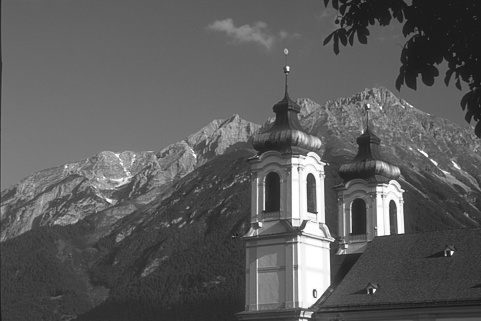}
}
\subfigure[{\scriptsize Noisy image. $L=1$}]{
\centering
\includegraphics[width=0.3\textwidth]{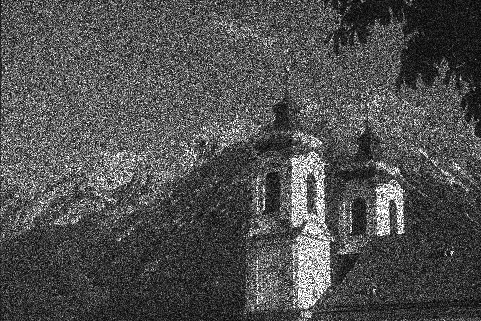}
}
\subfigure[{ \scriptsize $\Gamma$-Map (22.49/0.578/0.182)}]{
\centering
\includegraphics[width=0.3\textwidth]{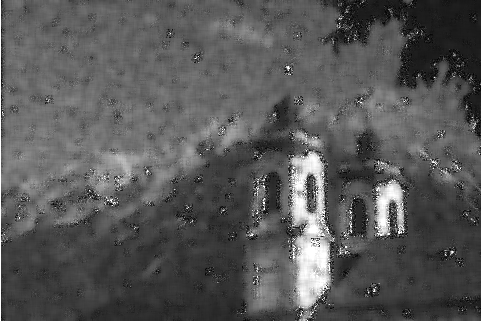}
}\\
\subfigure[{\scriptsize SRAD ( 23.47/0.728/0.300)}]{
\centering
\includegraphics[width=0.3\textwidth]{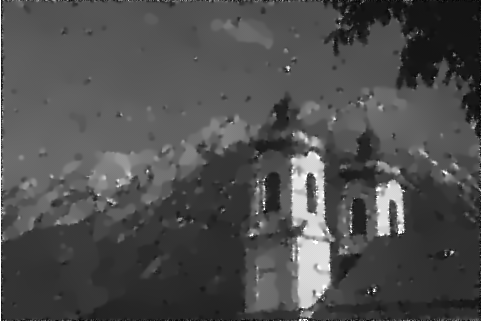}
}
\subfigure[{\scriptsize DPAD (23.54/0.729/0.318)}]{
\centering
\includegraphics[width=0.3\textwidth]{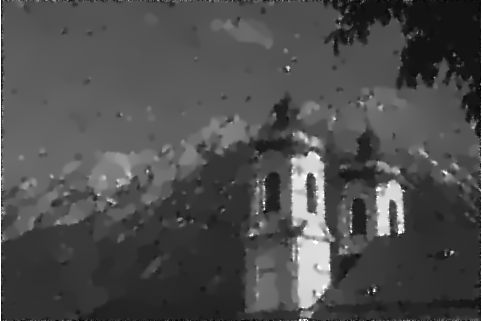}
}
\subfigure[{\scriptsize $\mathrm{TRDMD}_{7 \times 7}^{10}$} \scriptsize({\textbf{26.80}/\textbf{0.752}/\textbf{0.461}})]{
\centering
\includegraphics[width=0.3\textwidth]{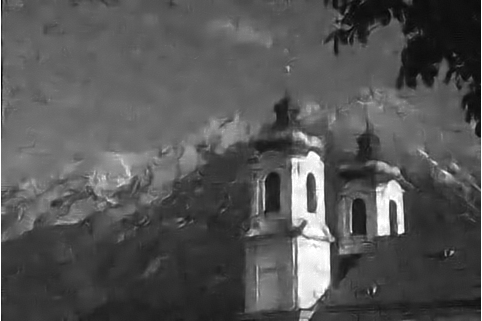}
}
\caption{Despeckling of $Mountain$ with $L=1$. The results are reported by PSNR/MSSIM/EC index. Best results are marked.}
\label{L1comp}
\end{figure*}

\subsection{Influence of Diffusion Stages}
In this study, any number of diffusion stages can be exploited in our model. But in practice, the trade-off between run time and accuracy should be considered. Therefore, we need to study the influence of the number of diffusion stages on the denoising performance. $\mathrm{TRDMD}_{5 \times 5}^T$ and 400 images are used for training.

As shown in Fig.~\ref{influence}(c), the performance improvement becomes quite insignificant when the diffusion stages $\geq 10$. In order to save the training time, we choose $\mathrm{diffusion  \; stage} = 10$ in the following experiments as it provides
the best trade-off between performance and computation time.

\begin{figure*}[t]
\centering
\subfigure[{\scriptsize$water$ $castle$}]{
\centering
\includegraphics[width=0.23\textwidth]{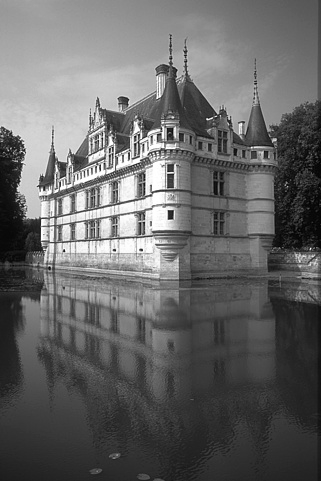}
}
\subfigure[\scriptsize{ Noisy image. $L=3$}]{
\centering
\includegraphics[width=0.23\textwidth]{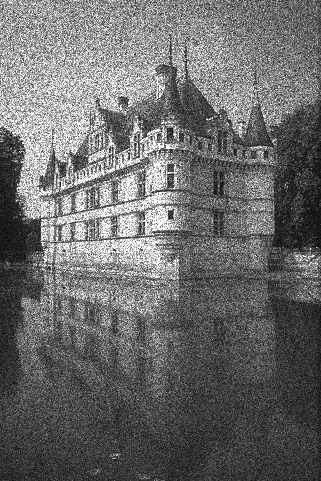}
}
\subfigure[{\scriptsize $\Gamma$-Map (23.27/0.687/0.420)}]{
\centering
\includegraphics[width=0.23\textwidth]{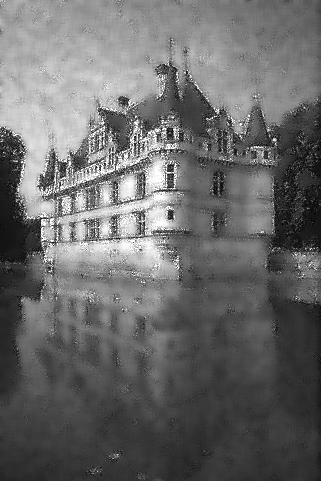}
}\\
\subfigure[{\scriptsize SRAD (25.38/0.799/0.525)}]{
\centering
\includegraphics[width=0.23\textwidth]{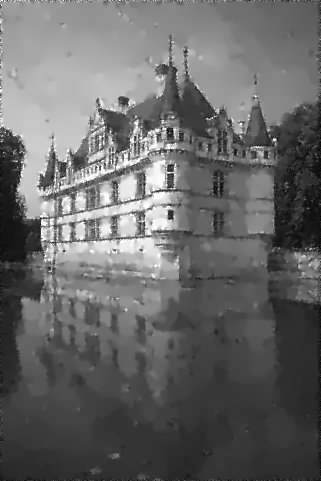}
}
\subfigure[{\scriptsize DPAD (25.45/0.799/0.532)}]{
\centering
\includegraphics[width=0.23\textwidth]{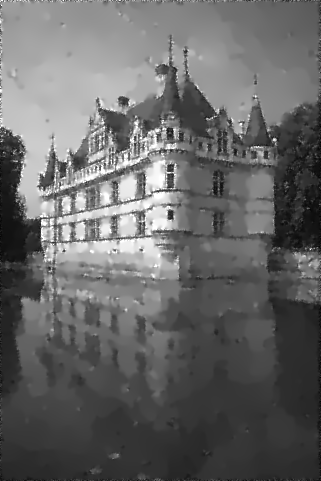}
}
\subfigure[{\scriptsize$\mathrm{TRDMD}_{7 \times 7}^{10}$ (\textbf{27.20}/\textbf{0.817}/\textbf{0.647})}]{
\centering
\includegraphics[width=0.23\textwidth]{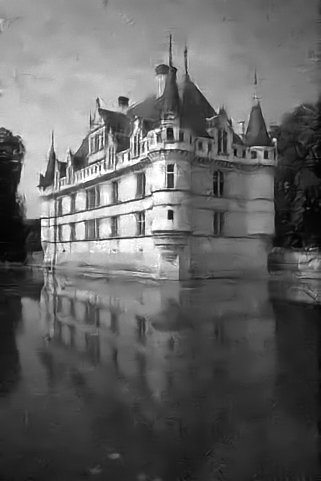}
}
\caption{Despeckling of $Tower$ with $L=3$. The results are reported by PSNR/MSSIM/EC index. Best results are marked.}
\label{L3comp}
\vspace{-0.5cm}
\end{figure*}

\subsection{Experimental Results}
By analyzing the above three subsections, we decide to employ $\mathrm{TRDMD}_{7 \times 7}^{10}$ model and 400 images for training.
Note that, the diffusion model needs to be trained respectively for different number of looks.

\subsubsection{Comparison with Traditional Anisotropic Diffusion Based Approaches}
\begin{table*}[t]
\scriptsize
\centering
\begin{tabular}{|c |c |c |c |c |}
\hline
 & $L$=1 & $L$=3 & $L$=5 &$L$=8\\
\hline
 & {\scriptsize PSNR/MSSIM/EC} & {\scriptsize PSNR/MSSIM/EC} & {\scriptsize PSNR/MSSIM/EC} &{\scriptsize PSNR/MSSIM/EC}\\
\hline
$\Gamma$-Map &{\scriptsize 21.02/0.513/0.238}&{\scriptsize 23.46/0.651/0.394}&{\scriptsize 24.70/0.708/0.469}&{\scriptsize 25.97/0.757/0.540 }\\
\hline
SRAD &{\scriptsize 24.82/0.686/0.465} &{\scriptsize 25.42/0.738/0.489}&{\scriptsize 26.92/0.790/0.570}&{\scriptsize 28.25/0.819/0.639} \\
\hline
DPAD &{\scriptsize 24.78/0.681/0.473} &{\scriptsize 25.48/0.735/0.499}&{\scriptsize 27.00/0.789/0.577}&{\scriptsize 28.29/0.818/0.643}\\
\hline
$\mathrm{TRDMD}_{7 \times 7}^{10}$& {\scriptsize \textbf{24.95}/\textbf{0.678}/\textbf{0.479}}&{\scriptsize\textbf{27.28}/\textbf{0.775}/\textbf{0.630}}&{\scriptsize\textbf{28.53}/\textbf{0.819}/\textbf{0.700}}
&{\scriptsize\textbf{29.70}/\textbf{0.852}/\textbf{0.754}}\\
\hline
\end{tabular}
\vspace*{0.2cm}
\caption{Comparison of the performance of the test algorithms in terms of PSNR, MSSIM and EC. Best results are marked.}
\label{resultshow0}
\end{table*}
In this subsection, we compare the proposed despeckling model with the $\Gamma$-Map filter \cite{lopes1990maximum} and some related anisotropic diffusion approaches, e.g., speckle
reducing anisotropic diffusion (SRAD) \cite{yu2002speckle} and detail preserving anisotropic diffusion
(DPAD) \cite{aja2006estimation}. Moreover, to ensure that SRAD and DPAD achieve their best performance respectively, we set a time step $\Delta t = 0.2$ and run 400-3000 iterations for different noise levels.

By observation on despeckling performances in Fig.~\ref{L1comp} and Fig.~\ref{L3comp}, we can see that in comparison with the $\Gamma$-Map filter, the other three methods show superior performance on edge localization and detail preservation. However, the unnatural smoothness introduced
after iterative processing makes SRAD and DPAD produce
cartoon-like images, i.e., made up by textureless geometric
patches. Therefore, SRAD and DPAD may be unsuitable for practical application,
because fine details and textures that may be useful for analysis are destroyed. On the contrary, the proposed method produces more natural results in visual quality and preserves more detailed information.

Quantitative evaluation results are shown in Table~\ref{resultshow0}. Overall speaking, DPAD gives slightly better results than SRAD in terms of evaluation indexes. Meanwhile, Table~\ref{resultshow0} show that the best indexes are provided by the proposed model, indicating that the proposed model is more powerful in structural and edge preservation.

\subsubsection{Comparison with State-of-the-art Despeckling Approaches}
In this subsection, we compare the proposed despeckling model with several state-of-the-art despeckling approaches, i.e., the nonlocal methods PPBit \cite{deledalle2009iterative}, SAR-BM3D \cite{ParrilliPAV12}, and our previous FoE-based despeckling work \cite{chen2014higher}.

\begin{figure*}[t]
\centering
\subfigure[\scriptsize{$Flower$}]{
\centering
\includegraphics[width=0.3\textwidth]{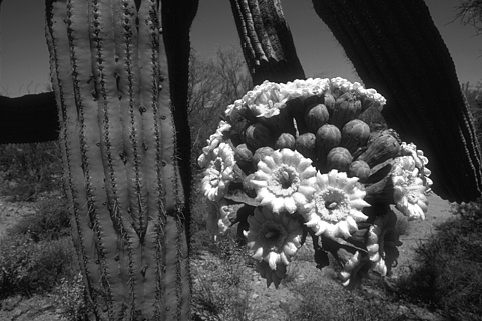}
}
\subfigure[\scriptsize{ Noisy image. $L=1$}]{
\centering
\includegraphics[width=0.3\textwidth]{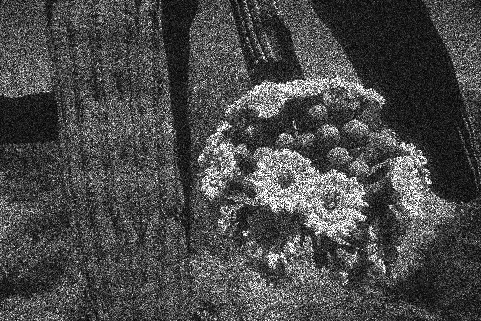}
}
\subfigure[\scriptsize{ PPBit (23.22/0.595/0.278)}]{
\centering
\includegraphics[width=0.3\textwidth]{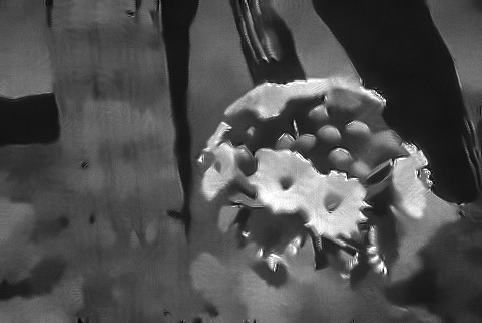}
}\\
\subfigure[\scriptsize{ SAR-BM3D (24.40/0.672/0.354)}]{
\centering
\includegraphics[width=0.3\textwidth]{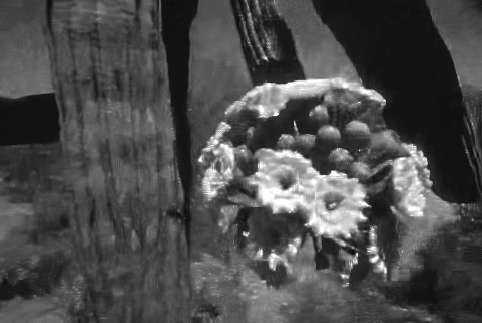}
}
\subfigure[\scriptsize{ FoE (24.08/0.637/0.388)}]{
\centering
\includegraphics[width=0.3\textwidth]{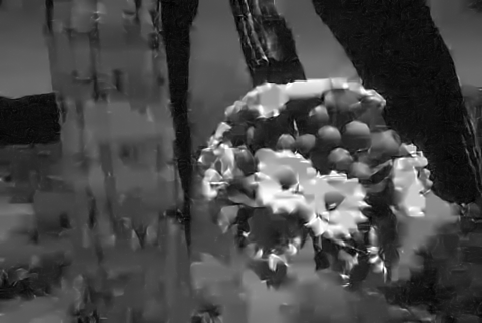}
}
\subfigure[\scriptsize{ $\mathrm{TRDMD}_{7 \times 7}^{10}$ (\textbf{24.53}/\textbf{0.675}/\textbf{0.392})}]{
\centering
\includegraphics[width=0.3\textwidth]{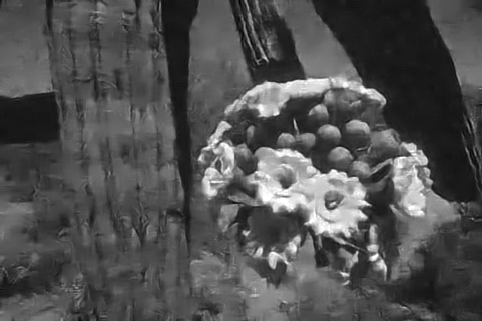}
}
\caption{Despeckling of $Flower$ with $L=1$. The results are reported by PSNR/MSSIM/EC index. Best results are marked.}
\label{L1syn}
\end{figure*}

\begin{figure*}[t]
\centering
\subfigure[\scriptsize{$Race$}]{
\centering
\includegraphics[width=0.3\textwidth]{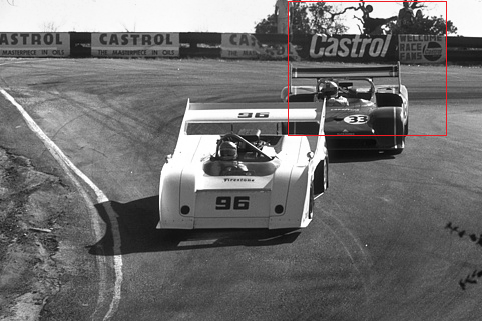}
}
\subfigure[\scriptsize{ Noisy image. $L=5$}]{
\centering
\includegraphics[width=0.3\textwidth]{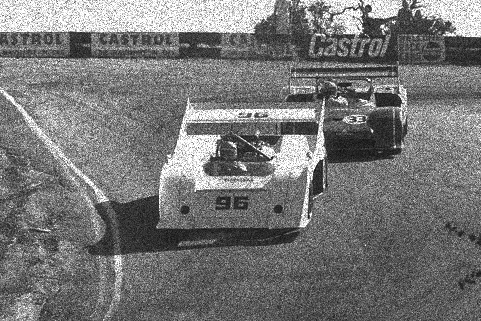}
}
\subfigure[\scriptsize{ PPBit (26.80/0.722/0.616)}]{
\centering
\includegraphics[width=0.3\textwidth]{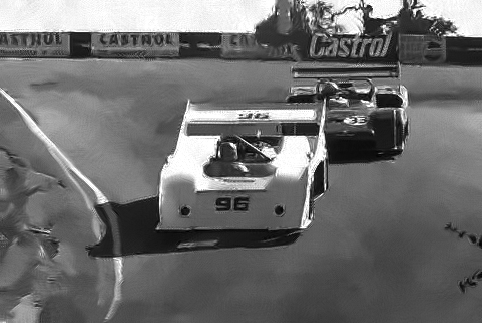}
}\\
\subfigure[\scriptsize{ SAR-BM3D (28.06/0.769/0.714)}]{
\centering
\includegraphics[width=0.3\textwidth]{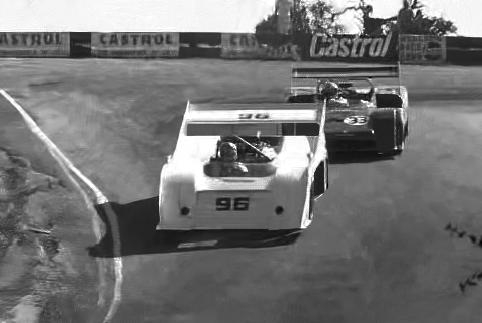}
}
\subfigure[\scriptsize{ FoE (27.94/0.755/0.745)}]{
\centering
\includegraphics[width=0.3\textwidth]{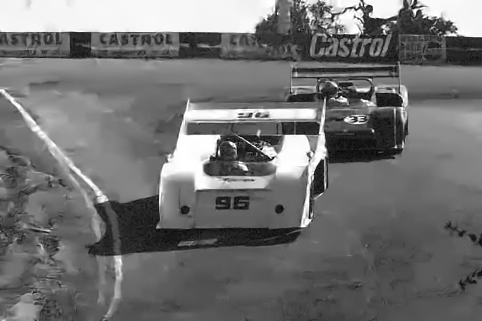}
}
\subfigure[\scriptsize{ $\mathrm{TRDMD}_{7 \times 7}^{10}$ (\textbf{28.36}/\textbf{0.772}/\textbf{0.751})}]{
\centering
\includegraphics[width=0.3\textwidth]{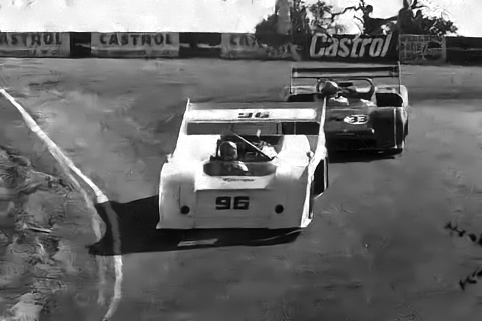}
}
\caption{Despeckling of $Race$ with $L=5$. The results are reported by PSNR/MSSIM/EC index. Best results are marked.}
\label{L5syn}
\end{figure*}

\begin{figure*}[t]
\centering
\subfigure[\scriptsize{ $Race$}]{
\centering
\includegraphics[width=0.25\textwidth]{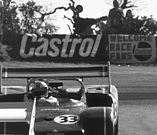}
}
\subfigure[\scriptsize{ Noisy image. $L=5$}]{
\centering
\includegraphics[width=0.25\textwidth]{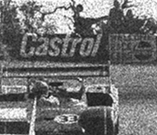}
}
\subfigure[\scriptsize{ PPBit}]{
\centering
\includegraphics[width=0.25\textwidth]{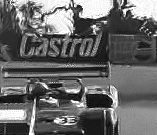}
}\\
\subfigure[\scriptsize{ SAR-BM3D}]{
\centering
\includegraphics[width=0.25\textwidth]{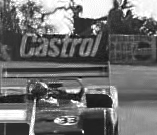}
}
\subfigure[\scriptsize{ FoE}]{
\centering
\includegraphics[width=0.25\textwidth]{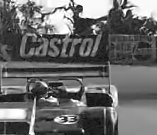}
}
\subfigure[\scriptsize{ $\mathrm{TRDMD}_{7 \times 7}^{10}$}]{
\centering
\includegraphics[width=0.25\textwidth]{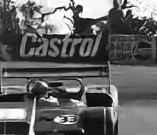}
}
\caption{The magnified despeckling results within the red rectangle in Fig.~\ref{L5syn}(a).}
\label{L5synzoom}
\end{figure*}

\begin{figure*}[t]
\centering
\subfigure[{ $Vancouver$. $L=1$}]{
\centering
\includegraphics[width=0.25\textwidth]{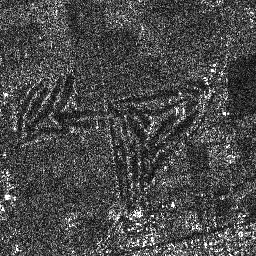}
}
\subfigure[{ $Bayard$. $L=1$}]{
\centering
\includegraphics[width=0.25\textwidth]{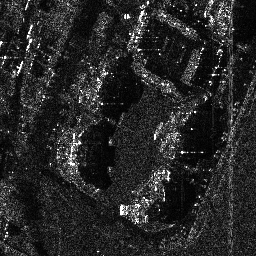}
}
\subfigure[{ $Cheminot$. $L=1$}]{
\centering
\includegraphics[width=0.25\textwidth]{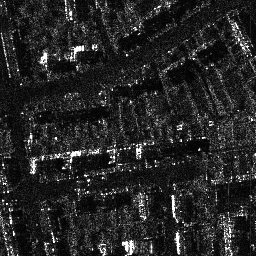}
}\\

\includegraphics[width=0.25\textwidth]{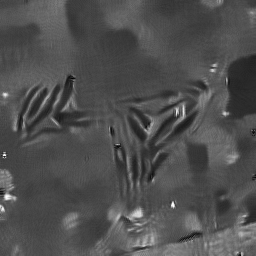}
\includegraphics[width=0.25\textwidth]{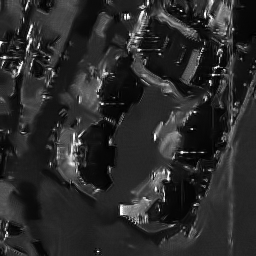}
\includegraphics[width=0.25\textwidth]{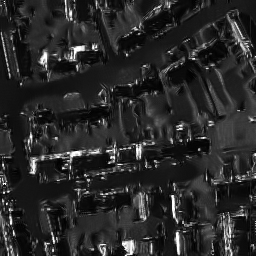}
\\
\vspace*{0.2cm}
\includegraphics[width=0.25\textwidth]{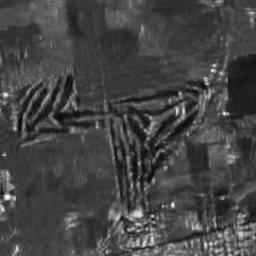}
\includegraphics[width=0.25\textwidth]{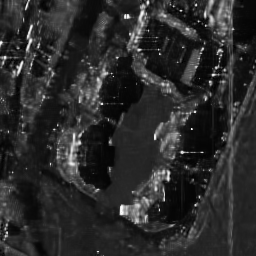}
\includegraphics[width=0.25\textwidth]{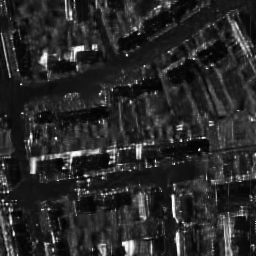}
\\
\vspace*{0.2cm}
\includegraphics[width=0.25\textwidth]{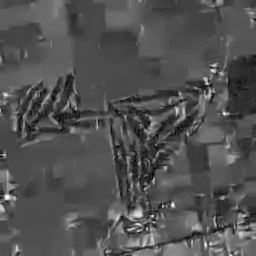}
\includegraphics[width=0.25\textwidth]{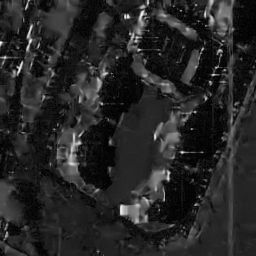}
\includegraphics[width=0.25\textwidth]{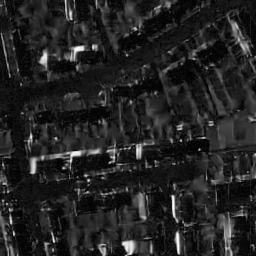}
\\
\vspace*{0.2cm}
\includegraphics[width=0.25\textwidth]{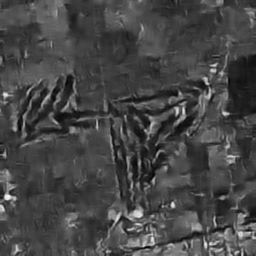}
\includegraphics[width=0.25\textwidth]{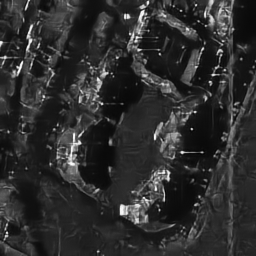}
\includegraphics[width=0.25\textwidth]{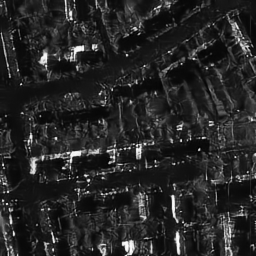}
\\
\caption{{ Despeckling of real SAR images with $L=1$. The employed methods are PPBit (the second row), SAR-BM3D (the third row), FoE (the fourth row) and the proposed $\mathrm{TRDMD}_{7 \times 7}^{10}$ (the fifth row).}}
\label{L1real}
\end{figure*}

Examining the recovery images with $L=1$ in Fig.~\ref{L1syn}, we see that in comparison with PPBit and FoE,
the proposed method and SAR-BM3D work better at capturing textures, details and small features. On the other hand, in comparison with SAR-BM3D,
the proposed method recovers clearer edges and is relatively less disturbed by artifacts. As shown in Fig.~\ref{L1syn}(d)-Fig.~\ref{L1real}(d)
the despeckled results
of SAR-BM3D are affected by the structured signal-like artifacts that appear in homogeneous areas of the image. This phenomenon is originated from
 the selection process in the BM3D denoising process. The selection process is easily influenced by the noise itself,
 especially in flat areas of the image, which can be dangerously self-referential.

\begin{table*}[t]
\scriptsize
\centering
\begin{tabular}{|c |c |c |c |c |}
\hline
 & $L$=1 & $L$=3 & $L$=5 &$L$=8\\
\hline
& {\scriptsize PSNR/MSSIM/EC} & {\scriptsize PSNR/MSSIM/EC} & {\scriptsize PSNR/MSSIM/EC} &{\scriptsize PSNR/MSSIM/EC}\\
\hline
Noisy &{\scriptsize 12.95/0.223/0.148} & {\scriptsize 17.42/0.378/0.255} &{\scriptsize 19.59/0.460/0.322}&{\scriptsize 21.61/0.535/0.393}\\
\hline
PPBit &{\scriptsize23.78/0.590/0.346} &{\scriptsize25.95/0.713/0.487 }&{\scriptsize26.95/0.763/0.549 }&{\scriptsize27.86/0.803/0.602} \\
\hline
SAR-BM3D &{\scriptsize24.85/0.676/0.455}&{\scriptsize27.12/\textbf{0.776}/0.605 }&{\scriptsize28.27/0.818/0.669}&{\scriptsize29.35/\textbf{0.852}/0.723} \\
\hline
FoE      &{\scriptsize24.65/0.669/\textbf{0.493}}&{\scriptsize27.07/0.763/\textbf{0.637}}& {\scriptsize28.32/0.812/0.698}&{\scriptsize29.54/0.848/0.753} \\
\hline
$\mathrm{TRDMD}_{7 \times 7}^{10}$& {\scriptsize \textbf{24.95}/\textbf{0.678}/0.479}&{\scriptsize\textbf{27.28}/0.775/0.630}&{\scriptsize\textbf{28.53}/\textbf{0.819}/\textbf{0.700}}
&{\scriptsize\textbf{29.70}/\textbf{0.852}/\textbf{0.754}}\\
\hline
\end{tabular}
\vspace*{0.2cm}
\caption{Comparison of the performance of the test algorithms in terms of PSNR, MSSIM and EC. Best results are marked.}
\label{resultshow}
\vspace*{-0.5cm}
\end{table*}
In Fig.~\ref{L5syn}-Fig.~\ref{L5synzoom} are reported the recovered results for $L=5$. For better visual comparison, we also provide the
magnified results within the red rectangles shown in Fig.~\ref{L5syn}.
By observation, one can see that the performance of $\mathrm{TRDMD}_{7 \times 7}^{10}$ is better than FoE and SAR-BM3D on detail preservation.
 In the visual quality, the typical structured artifacts encountered with the BM3D-based algorithm do not appear when the
 proposed method $\mathrm{TRDMD}_{7 \times 7}^{10}$ is used. Moreover, our method works better in geometry-preserving than FoE, which can be
 visually perceived by comparison on (e) and (f) in Fig.~\ref{L5synzoom}. Actually, one can see that functional \eqref{foemodel}
 is exactly the fields of experts (FoE) image prior regularized variational model for image restoration. However, in the $\mathrm{TRDMD}_{7 \times 7}^{10}$ model, both the linear filters and influence functions are trained and optimized, which is the critical factor for the effectiveness of the
 proposed diffusion despeckling model. This critical factor of the optimized diffusion model is quite different from the FoE prior based
variational model and traditional convolutional networks, where only linear filters are trained with fixed influence functions.

The recovery error in terms of PSNR (in dB) and MSSIM are summarized in Table~\ref{resultshow}. Comparing the
indexes in Table~\ref{resultshow}, the
overall performance of $\mathrm{TRDMD}_{7 \times 7}^{10}$ in terms of PSNR/MSSIM is better than the other methods.
This indicates that for most images our method is powerful in the recover quality and geometry feature preservation. Overall speaking, the proposed model and SAR-BM3D perform better than the other test methods in terms of texture and detail preservation, as illustrated in the MSSIM index. In addition, according to the EC index, the proposed model preserves more clearer edges in comparison with SAR-BM3D.

\begin{figure*}[t]
\centering
\subfigure[{ $Cropland$. $L=6$}]{
\centering
\includegraphics[width=0.25\textwidth]{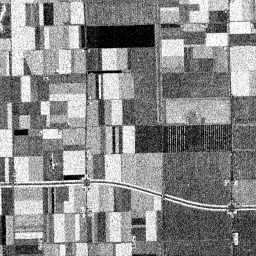}
}
\subfigure[{ PPBit}]{
\centering
\includegraphics[width=0.25\textwidth]{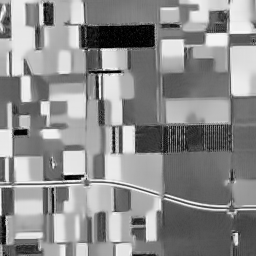}
}\\
\subfigure[{ SAR-BM3D}]{
\centering
\includegraphics[width=0.25\textwidth]{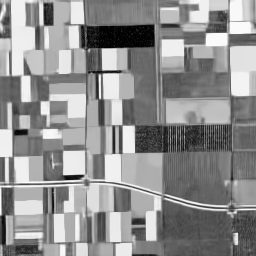}
}
\subfigure[{ FoE}]{
\centering
\includegraphics[width=0.25\textwidth]{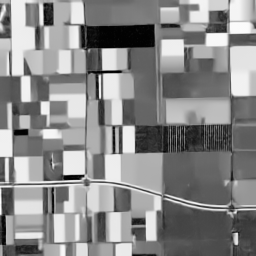}
}
\subfigure[{ $\mathrm{TRDMD}_{7 \times 7}^{10}$}]{
\centering
\includegraphics[width=0.25\textwidth]{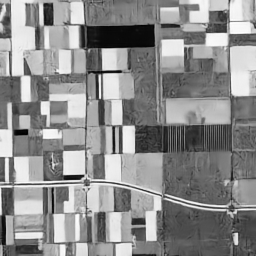}
}
\caption{Despeckling of $Cropland$ with $L=6$.}
\label{L3real}
\end{figure*}

We also present the despeckling results for real SAR amplitude images in Fig.~\ref{L1real}-Fig.~\ref{L3real}. The test four SAR images are as follows: 1)
single-look Radarsat-1 image of Vancouver (Canada) in Fig.~\ref{L1real}, identified as $Vancouver$; 2) single-look SAR images identified as $Bayard$ and $Cheminot$ from Saint-Pol-sur-Mer (France), sensed in 1996 by
RAMSES of ONERA, as shown in Fig.~\ref{L1real}; and 3) six-look AirSAR cropland scene identified as $Cropland$, as shown in Fig.~\ref{L3real}. By closely visual comparison, we can
observe that $\mathrm{TRDMD}_{7 \times 7}^{10}$ recovers clearer texture and sharper edges. Especially for the tiny features,
 $\mathrm{TRDMD}_{7 \times 7}^{10}$ is able to catch but the other methods neglects them. Although
these features are not quite obvious, the trained diffusion model still extracts them and exhibit these features apparently.

Table~\ref{resultreal} lists the three without-reference indexes of the different
algorithms on four real SAR images. Comparing the
indexes in Table~\ref{resultreal}, one can see that no algorithm is predominant in the despeckling performance. However, overall speaking, the proposed model provides most preferable results in terms of the evaluation indexes. Overall speaking, the proposed model provides comparable or even better results than the test state-of-the-art methods, both in visual effects and the numerical indexes.
\begin{table}[t!]
\scriptsize
\centering
\begin{tabular}{|c |c |c |c |c |c |c |}
\hline
{\footnotesize Image} &{\footnotesize Index}&{\footnotesize Ideal}& {\footnotesize PPBit} &{\footnotesize BM3D }&{\footnotesize FoE }& {\footnotesize$\mathrm{TRDMD}$} \\
\hline
 &{\footnotesize RI$\_$M}&{\footnotesize 1}&{\footnotesize 0.862}&{\footnotesize 0.873}&{\footnotesize \textbf{0.913}}&{\footnotesize 0.900}\\
 {\footnotesize$Vancouver$}&{\footnotesize RI$\_$V}&{\footnotesize 0.273}&{\footnotesize \textbf{0.198}}&{\footnotesize 0.179 }&{\footnotesize 0.195}&{\footnotesize 0.179}\\
 &{\footnotesize $C_u$}&{\footnotesize 0.304} &{\footnotesize 0.236}&{\footnotesize 0.284}&{\footnotesize 0.281}&{\footnotesize \textbf{0.314}}\\
\hline
 &{\footnotesize RI$\_$M}&{\footnotesize1}&{\footnotesize0.861}&{\footnotesize0.850}&{\footnotesize\textbf{1.020}}&{\footnotesize0.859} \\
 {\footnotesize$Bayard$}&{\footnotesize RI$\_$V}&{\footnotesize0.273}&{\footnotesize0.178}&{\footnotesize0.157}&{\footnotesize0.073}&{\footnotesize\textbf{0.203}}\\
 &{\footnotesize$C_u$}&{\footnotesize0.770}&{\footnotesize0.744}&{\footnotesize0.736}&{\footnotesize0.838}&{\footnotesize\textbf{0.750}}  \\
\hline
 &{\footnotesize RI$\_$M}&{\footnotesize1}&{\footnotesize0.863}&{\footnotesize0.837}&{\footnotesize\textbf{1.027}}&{\footnotesize0.847}  \\
 {\footnotesize$Cheminot$} & {\footnotesize RI$\_$V}&{\footnotesize0.273}&{\footnotesize 0.165}&{\footnotesize0.147}&{\footnotesize 0.060}&{\footnotesize\textbf{0.203}}\\
 &{\footnotesize $C_u$}&{\footnotesize 0.844}&{\footnotesize \textbf{0.848}}&{\footnotesize0.808}&{\footnotesize0.932}&{\footnotesize0.829} \\
\hline
 &{\footnotesize RI$\_$M}&{\footnotesize1}&{\footnotesize\textbf{0.974}}&{\footnotesize0.969}&{\footnotesize0.965}&{\footnotesize0.973} \\
 {\footnotesize$Cropland$}&{\footnotesize RI$\_$V}&{\footnotesize0.046}&{\footnotesize0.037}&{\footnotesize 0.038}&{\footnotesize 0.028}& {\footnotesize\textbf{0.048}} \\
 &{\footnotesize$C_u$}&{\footnotesize0.337}&{\footnotesize\textbf{0.412}}&{\footnotesize0.415}&{\footnotesize0.426}&{\footnotesize0.423}  \\
\hline
\end{tabular}
\vspace*{0.2cm}
\caption{Comparison of the performance of the test algorithms on real SAR images. Best results are marked. Note that in the table BM3D refers to SAR-BM3D method and $\mathrm{TRDMD}$ denotes $\mathrm{TRDMD}_{7 \times 7}^{10}$.}
\label{resultreal}
\vspace*{-0.5cm}
\end{table}
\subsubsection{Run Time}

The proposed method merely contains
convolution of linear filters with an image, which offers high computation efficiency and high levels of parallelism
making it well suited for GPU implementation.
More precisely, the proposed model can not only run fast on CPU especially when the 2-D convolution is efficiently realized by FFT, but also be implemented using GPU. Even without GPU, the proposed model is efficient enough.

In Table \ref{runtime}, we report the typical run time of our model
for the images of two different dimensions for the case of $L = 8$. It is worthy noting that the FoE-based method needs more iterations to converge if the noise level is higher. Therefore, for $L=1, 3$ or 5, the consuming time of FoE-based method is more than $L=8$.
We also present the run time of four competing algorithms for a comparison. Note that the method FANS \cite{CozzolinoPSPV14} is the improved version of SAR-BM3D,
at the expense of slight performance degradation. Meanwhile, the main body of SAR-BM3D, PPBit and FANS are all implemented in C language.
All the methods are run in Matlab with single-threaded computation for CPU implementation.

Due to the structural simplicity of our model, it is well-suited to GPU parallel computation. We are able to implement our algorithm on
GPU with ease. It turns out that the GPU implementation
based on NVIDIA Geforce GTX 780Ti can accelerate the inference procedure significantly, as shown in Table \ref{runtime}. By comparison, we
see that our $\mathrm{TRDMD}_{7 \times 7}^{10}$ model is generally faster than the other methods, especially with GPU implementation. This is reasonable because the PPBit and SAR-BM3D are based nonlocal approach with high computation complexity.
For FoE-based method, the required maximum number of iterations is 120 for $L=8$, 150 for $L=5$, 300 for $L=3$ and 600 for $L=1$. On the contrary, our $\mathrm{TRDMD}_{7 \times 7}^{10}$ model needs only 10 iteration steps with similar amount of computations at each step in comparison with FoE.

\begin{table}[t!]
\begin{center}
\begin{tabular}{r|c|c|c|c|c}
\cline{1-6}
& SAR-BM3D & FANS& PPBit & FoE ($L$=8)&$\mathrm{TRDMD}_{7 \times 7}^{10}$\\
\hline\hline
$256 \times 256$ & 42.4 & 3.71& 13.2 &  14.86(0.12)&3.05 (\textbf{0.03}) \\
$512 \times 512$ & 169.1 &14.40 & 48.9 &59.43(0.37)&9.33 (\textbf{0.09})\\
\cline{1-6}
\end{tabular}
\end{center}
\caption{Typical run time (in second) of the Despeckling methods for images with two different dimensions.
The CPU computation time is evaluated on Intel CPU X5675, 3.07GHz.
The highlighted number is the run time of GPU implementation based on NVIDIA Geforce GTX 780Ti.}
\label{runtime}
\vspace*{-0.5cm}
\end{table}

\section{Conclusion}
In this study we proposed a simple but effective despeckling approach with both high computational efficiency and comparable
  or even better results than state-of-the-art approaches. We achieve this goal by exploiting the newly-developed trainable nonlinear reaction diffusion model. The linear filters and influence functions in the model are simultaneously optimized by taking into account the speckle noise statistics.

The proposed model merely contains
convolution of linear filters with an image, which offers high computation efficiency on CPU. Moreover, high levels of parallelism in the proposed model
make it well suited for GPU implementation. Therefore, the proposed model is able to deal with massive-scale and huge amount of data.
In comparison with several related anisotropic diffusion despeckling approaches with handcrafted filters and unique influence function, e.g., SRAD and DPAD, the proposed model provides a much enhanced performance in both visual effects and evaluation indexes. In comparison with the nonlocal-based PPBit and SAR-BM3D methods, the proposed model bears the properties of simple structure
  and high efficiency with strongly competitive despeckling performance.

  Note that, the training data is our study is natural images. Hence, the despeckling performance on real SAR images could be enhanced if the remote sensing dataset is employed for training.
{
\bibliographystyle{ieee}
\bibliography{references}
}

\end{document}